\def\paperID{9289}
\def\confName{CVPR}
\def\confYear{2024}

\def\paperTitle{Entangled View-Epipolar Information Aggregation for \\ Generalizable Neural Radiance Fields}

\def\authorBlock{
    Zhiyuan Min\textsuperscript{1}  \qquad
    Yawei Luo\textsuperscript{1,}\thanks{Corresponding author} \qquad
    Wei Yang\textsuperscript{2} \qquad
    Yuesong Wang\textsuperscript{2} \qquad
    Yi Yang\textsuperscript{1} \qquad \\
    \textsuperscript{1}Zhejiang University, China \qquad
    \textsuperscript{2}Huazhong University of Science and Technology, China \\
    {\tt\small \{minzhiyuan, yaweiluo, yangyics\}@zju.edu.cn} \qquad
    {\tt\small \{weiyangcs, yuesongwang\}@hust.edu.cn}
}

\newif\ifreview 
\newif\ifarxiv \newcommand{\arxiv}{\arxivtrue}
\newif\ifcamera 
\newif\ifrebuttal

\arxiv 

\pdfoutput=1
\documentclass[10pt,twocolumn,letterpaper]{article}
\ifreview \usepackage[review]{cvpr} \fi
\ifarxiv \usepackage[pagenumbers]{cvpr} \fi
\ifrebuttal \usepackage[rebuttal]{cvpr} \fi
\ifcamera \usepackage{cvpr} \fi

\usepackage{graphicx}	
\usepackage{amsmath}	
\usepackage{amssymb}	
\usepackage{booktabs}
\usepackage{times}
\usepackage{microtype}
\usepackage{epsfig}
\usepackage[table,xcdraw,dvipsnames]{xcolor}
\usepackage{caption}
\usepackage{float}
\usepackage{placeins}
\usepackage{color, colortbl}
\usepackage{stfloats}
\usepackage{enumitem}
\usepackage{tabularx}
\usepackage{xstring}
\usepackage{multirow}
\usepackage{xspace}
\usepackage{url}
\usepackage{subcaption}
\usepackage{xcolor}
\usepackage[hang,flushmargin]{footmisc}

\ifcamera \usepackage[accsupp]{axessibility} \fi

% ...

\ifarxiv  \fi

\newcommand{\R}[1]{{%
    \textbf{%
        \ifstrequal{#1}{1}{\textcolor{red}{R#1}}{%
        \ifstrequal{#1}{2}{\textcolor{blue}{R#1}}{%
        \ifstrequal{#1}{3}{\textcolor{magenta}{R#1}}{%
        \ifstrequal{#1}{4}{\textcolor{teal}{R#1}}{%
                           \textcolor{cyan}{R#1}%
        }}}}%
    }%
}}

\usepackage{xr-hyper}

\makeatletter
\newcommand*{\addFileDependency}[1]{
  \typeout{(#1)}
  \@addtofilelist{#1}
  \IfFileExists{#1}{}{\typeout{No file #1.}}
}

\makeatother

\definecolor{cvprblue}{rgb}{0.21,0.49,0.74}
\usepackage[pagebackref,breaklinks,colorlinks,citecolor=cvprblue]{hyperref}
\usepackage[capitalize]{cleveref}
\usepackage[ruled,linesnumbered]{algorithm2e}
\usepackage{wrapfig}
\crefname{section}{Sec.}{Secs.}
\crefname{table}{Table}{Tables}
\crefname{figure}{Fig.}{Figs.}

\begin{document}
\title{\paperTitle}
\author{\authorBlock}
\maketitle

\begin{abstract}

Generalizable NeRF can directly synthesize novel views across new scenes, eliminating the need for scene-specific re-training in vanilla NeRF. A critical enabling factor in these approaches is the extraction of a generalizable 3D representation by aggregating source-view 
features. In this paper, we propose an Entangled View-Epipolar Information Aggregation method dubbed \textbf{\emph{EVE-NeRF}}. Different from existing methods that consider cross-view and along-epipolar information independently, EVE-NeRF conducts the view-epipolar feature aggregation in an entangled manner by injecting the scene-invariant appearance continuity and geometry consistency priors to the aggregation process. Our approach effectively mitigates the potential lack of inherent geometric and appearance constraints resulting from one-dimensional interactions, thus further boosting the 3D representation generalizability. EVE-NeRF attains state-of-the-art performance across various evaluation scenarios. Extensive experiments demonstrate that, compared to prevailing single-dimensional aggregation, the entangled network excels in the accuracy of 3D scene geometry and appearance reconstruction. Our code is publicly available at  \href{https://github.com/tatakai1/EVENeRF}{https://github.com/tatakai1/EVENeRF}.

\end{abstract}
\section{Introduction}
\label{sec:intro}

The neural radiance fields (NeRF)~\cite{mildenhall2021nerf}, along with its subsequent refinements~\cite{barron2021mip, yu2021plenoctrees, barron2022mip}, have exhibited remarkable efficacy in the domain of novel view synthesis~\cite{luo2019pmvsnet,luo2020attention,wang2020mesh,wang2023adaptive}, showcasing immense potential for applications in smart education~\cite{luo2024LDMC,ma2023personas}, medical treatment~\cite{guan2021discriminative,molaei2023implicit}, and automatic driving~\cite{luo2019taking,luo2019significance,luo2021category,luo2020adversarial_NIPS2020}.  Despite these advancements, the methods along this vein often pertain to the training scene thus necessitating re-training for synthesizing new scenes. Such drawbacks severely constrain their practical applications.

More recently, the development of generalizable NeRF models~\cite{yu2021pixelnerf,wang2021ibrnet,chen2021mvsnerf} has emerged as a promising solution to address this challenge. These models can directly synthesize novel views across new scenes, eliminating the need for scene-specific re-training. A critical enabling factor in these approaches is the synthesis of a generalizable 3D representation by aggregating source-view features. Instead of densely aggregating every pixel in the source images, prior works draw inspiration from the epipolar geometric constraint across multiple views to aggregate view or epipolar information~\cite{johari2022geonerf,reizenstein2021common, varma2022attention, suhail2022generalizable, wang2022generalizable}. To capitalize on cross-view prior, specific methodologies~\cite{varma2022attention, johari2022geonerf} interact with the re-projected feature information in the reference view at a predefined depth. On the along-epipolar aspect~\cite{suhail2022generalizable, suhail2022light}, some methods employ self-attention mechanisms to sequentially obtain the entire epipolar line features in each reference view. 

We posit that both view and epipolar aggregation are crucial for learning a generalizable 3D representation: cross-view feature aggregation is pivotal to capturing geometric information, as the features from different views that match tend to be on the surface of objects. Concurrently, epipolar feature aggregation contributes by extracting depth-relevant appearance features from the reference views associated with the target ray, thus achieving a more continuous appearance representation. Nevertheless, the prevailing methods often execute view and epipolar aggregation independently~\cite{varma2022attention, johari2022geonerf} or in a sequential manner~\cite{suhail2022generalizable}, thereby overlooking the simultaneous interaction of appearance and geometry information.

In this paper, we introduce a novel \textbf{E}ntangled \textbf{V}iew-\textbf{E}pipolar information aggregation network, denoted as EVE-NeRF. EVE-NeRF is designed to enhance the quality of generalizable 3D representation through the simultaneous utilization of complementary appearance and geometry information. The pivotal components of EVE-NeRF are the View-Epipolar Interaction Module (VEI) and the Epipolar-View Interaction Module (EVI). Both modules adopt a dual-branch structure to concurrently integrate view and epipolar information. On one hand, VEI comprises a view transformer in its first branch to engage with the features of sampling points re-projected on all source views. In the second branch, VEI is equipped with an Along-Epipolar Perception submodule to inject the appearance continuity prior to the view aggregation results. On the other hand, EVI consists of an epipolar transformer in its first branch to aggregate features from sampling points along the entire epipolar line in each source view. In the second branch, EVI utilizes a Multi-View Calibration submodule to incorporate the geometry consistency prior to the epipolar aggregation representation. The alternating organization of EVI and VEI results in a generalizable condition for predicting the color of target rays based on NeRF volumetric rendering. 

Compared to the prevailing methods such as GNT~\cite{varma2022attention} and GPNR~\cite{suhail2022generalizable}, EVE-NeRF distinguishes itself in its ability to synthesize a target ray by entangling epipolar and view information. This capability serves to offset the appearance and geometry prior losses that typically arise from single-dimensional aggregation operations (see Figure \ref{intro_fig1}). Our main contributions can be summarized as follows:\vspace{5pt}
\begin{itemize}
    \item Through extensive investigation, we have revealed the under-explored issues of prevailing cross-view and along-epipolar information aggregation methods for generalizable NeRF. 
    \item We propose EVE-NeRF, which harnesses the along-epipolar and cross-view information in an entangled manner. EVE-NeRF complements the cross-view aggregation with appearance continuity prior and calibrates the along-epipolar aggregation with geometry consistency prior.
    \item EVE-NeRF produces more realistic novel-perspective images and depth maps for previously unseen scenes without any additional ground-truth 3D data. Experiments demonstrate that EVE-NeRF achieves state-of-the-art performance in various novel scene synthesis tasks.
\end{itemize}

\section{Related Work}
\label{sec:related}

\textbf{NeRF and Generalizable NeRF.}\; Recently, NeRF~\cite{mildenhall2021nerf} has made groundbreaking advancements in the field of novel view synthesis through a compact implicit representation based on differentiable rendering. Subsequent developments in the NeRF framework have explored various avenues, enhancing rendering quality~\cite{zhang2020nerf++,barron2021mip,barron2022mip,hu2023tri,barron2023zip}, accelerating rendering speed~\cite{yu2021plenoctrees,hedman2021baking, sun2022direct, chen2022tensorf,reiser2023merf, chen2023mobilenerf}, applicability to both rigid and non-rigid dynamic scenes~\cite{park2021nerfies,pumarola2021d,tretschk2021non, tian2023mononerf,gao2021dynamic,zheng2022structured}, and extending its capabilities for editing~\cite{liu2021editing, wang2022clip,sun2022fenerf,xu2022deforming, xiang2021neutex}.

The original NeRF and its subsequent improvements have achieved successful performance but suffered from the limitation of being trainable and renderable only in a single scene, which restricts their practical applications~\cite{feng2023clustering,yin2022proposalcontrast,yin2022semi}. One solution to this issue is conditioning on CNN features from the known view images, which align with the input coordinates of NeRF.
PixelNeRF~\cite{yu2021pixelnerf} encodes input images into pixel-aligned feature grids, combining image features with corresponding spatial positions and view directions in a shared MLP to output colors and densities. MVSNeRF~\cite{chen2021mvsnerf} utilizes a cost volume to model the scene, with interpolated features on volume conditioned.
Our approach also employs CNN features from known views, and we input the processed, pixel-aligned features into NeRF's MLP network to predict colors and densities. However, unlike PixelNeRF and similar methods~\cite{yu2021pixelnerf, chen2021mvsnerf,chen2023explicit}, which use average pooling for handling multiple views, our approach learns multi-view information and assigns weights to each view based on its relevance.\vspace{8pt}

\noindent\textbf{Generalizable NeRF with Transformers.}\; More recently, generalizable novel view synthesis methods~\cite{wang2021ibrnet,suhail2022generalizable,kulhanek2022viewformer,wang2022generalizable,chibane2021stereo,li2021mine,reizenstein2021common, huang2023local, cong2023enhancing} have incorporated transformer-based networks to enhance visual features from known views. These approaches employ self-attention or cross-attention mechanisms along various dimensions such as depth, view, or epipolar, enabling high-quality feature interaction and aggregation.
GeoNeRF~\cite{johari2022geonerf} concatenates view-independent tokens with view-dependent tokens and feeds them into a cross-view aggregation transformer network to enhance the features of the cost volume. GPNR~\cite{suhail2022generalizable} employs a 3-stage transformer-based aggregation network that sequentially interacts with view, epipolar, and view information. GNT~\cite{varma2022attention} and its subsequent work, GNT-MOVE~\cite{cong2023enhancing}, utilize a 2-stage transformer-based aggregation network, first performing cross-view aggregation and then engaging depth information interaction. ContraNeRF~\cite{yang2023contranerf} initially employs a two-stage transformer-based network for geometry-aware feature extraction, followed by the computation of positive and negative sample contrastive loss based on ground-truth depth values.

Inspired by these developments, we have analyzed the limitations of single-dimensional aggregation transformer networks and introduced EVE-NeRF that achieves efficient interaction between the complementary appearance and geometry information across different dimensions. Moreover, our method does not require any ground-truth depth information for model training.\vspace{8pt}



\section{Problem Formulation}
\label{sec:problem formulation}
 \begin{figure*}
    \centering
    \includegraphics[width=1.0\textwidth]{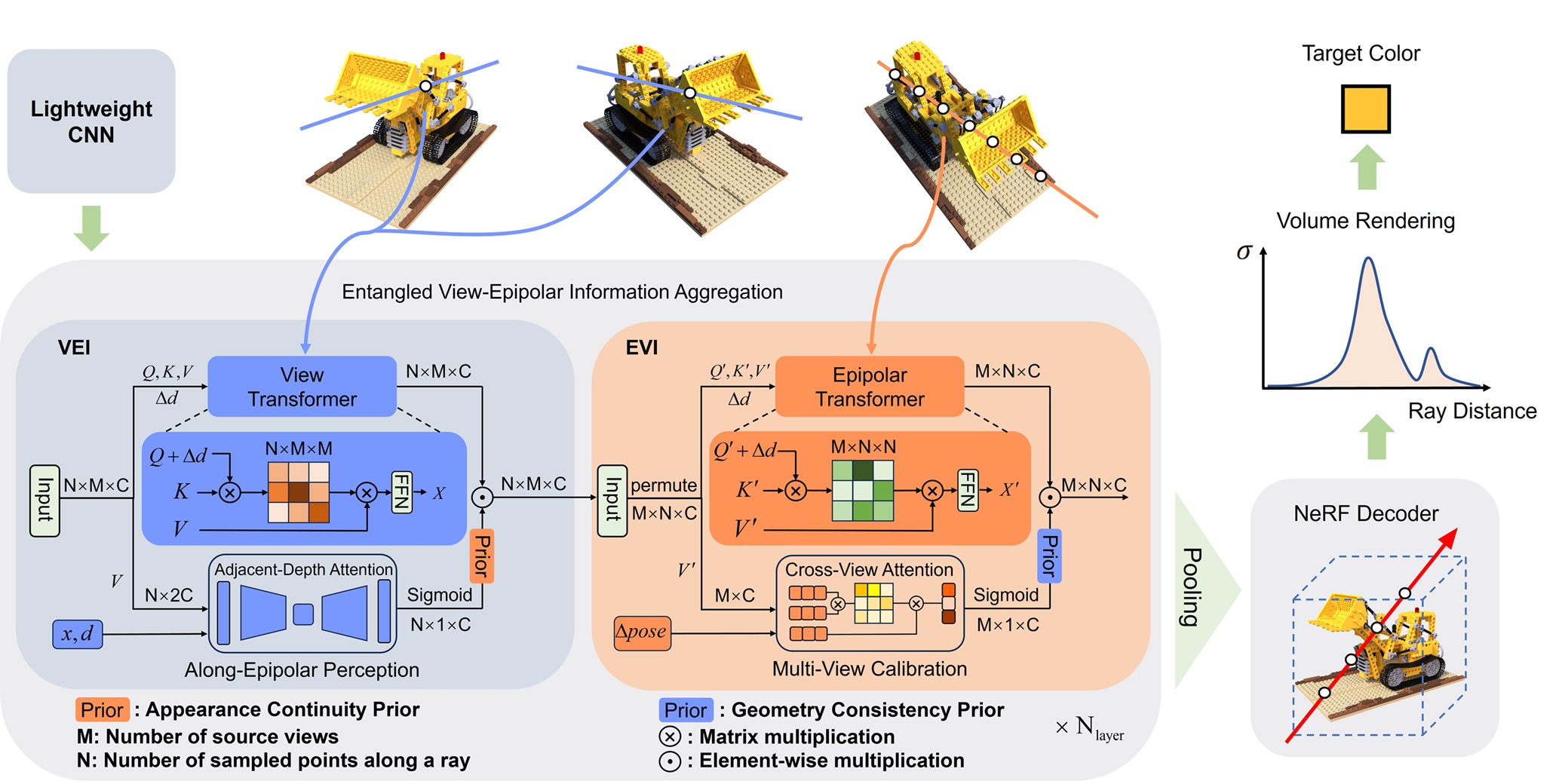}
    \caption{Pipline of EVE-NeRF. 1) We first employ a lightweight CNN to extract features of the epipolar sampling points from source views. 2) Through the Entangled View-Epipolar Information Aggregation, we complementarily enable information interaction in both the view and epipolar dimensions to produce generalizable multi-view epipolar features. 3) We use the NeRF Decoder to obtain color and density for the sampling points and predict the target color based on volume rendering.}
    \label{pipline1}
\end{figure*}


Our objective is to train a generalizable NeRF capable of comprehending 3D information from scenes the model has never encountered before and rendering new perspective images. 
Specifically, given $M$ source images for a particular scene $\boldsymbol{I}=\{\boldsymbol{I}_{i}\}_{i=1}^{M}$ and their corresponding camera intrinsic and extrinsic parameters $\boldsymbol{K}=\{ \boldsymbol{K}_{i}\}_{i=1}^{M}, \boldsymbol{P}=\{ \boldsymbol{P}_{i}=[\boldsymbol{R}_{i}, \boldsymbol{t}_{i}]\}_{i=1}^{M}$, most generalizable NeRF methods~\cite{yu2021pixelnerf, wang2021ibrnet,chen2021mvsnerf} can be formulated with a generalizable feature extraction network $\mathcal F_{\theta}$ and a rendering network $\mathcal G_{\phi}$:
\begin{gather}
    \mathcal F_{\theta}:(\boldsymbol{I}, \boldsymbol{K}, \boldsymbol{P}) \rightarrow \boldsymbol{z}, \quad \mathcal G_{\phi}:(\boldsymbol{x}, \boldsymbol{d}, \boldsymbol{z}) \rightarrow (\boldsymbol{c}, \sigma),
    \label{formula1}
\end{gather}
where $\boldsymbol{x}$ and $\boldsymbol{d}$ represent the 3D point position and the direction of the target ray's sampling points, while $\boldsymbol{c}$ and $\sigma$ are the predicted color and density, respectively. Similar to vanilla NeRF, $\boldsymbol{c}$ and $\sigma$ are utilized to compute the final color value of the target ray through volume rendering.
The variable $\boldsymbol{z}$ represents generalizable 3D representation of the scene provided by the feature extraction network. $\Theta$ and $\phi$ denote the learnable parameters of the networks.

\section{Methodology}
\label{sec:method}
\textbf{Overview.}\; Figure \ref{pipline1} provides an overview of EVE-NeRF, which includes a lightweight CNN-based image feature extractor, two dual-branch transformer-based modules named View-Epipolar Interaction (VEI) and Epipolar-View Interaction (EVI), respectively, and a conditioned NeRF decoder. Source images are first forwarded to a lightweight CNN and are transferred to feature maps. In the following, VEI and EVI are alternatively organized to aggregate the view-epipolar features in an entangled manner. The inter-branch information interaction mechanism within VEI and EVI capitalizes on the scene-invariant geometry and appearance priors to further calibrate the aggregated features. The output of the Entangle View-Epipolar Information Aggregation is a generalizable 3D representation $\boldsymbol{z}$. Finally, a conditioned NeRF decoder is employed for predicting the color and density values of the target ray based on $\boldsymbol{z}$ for volume rendering.

\subsection{Lightweight CNN}
For $M$ source views input $\{\boldsymbol{I}_{i}\}_{i=1}^{M}$, we first extract convolutional features $\{\boldsymbol{F}_{i}^{c}\}_{i=1}^{M}$ for each view independently using a lightweight CNN with sharing weights (see Appendix \ref{Additional Technical Details}). Unlike previous generalizable NeRF methods~\cite{chen2021mvsnerf, varma2022attention} that employ deep convolutional networks like U-Net~\cite{ronneberger2015u}, we use this approach since convolutional features with large receptive fields may not be advantageous for extracting scene-generalizable features~\cite{suhail2022generalizable}. Additionally, features derived from the re-projected sampling points guided by epipolar geometry are more focused on local information~\cite{huang2023local}. 

\subsection{View-Epipolar Interaction}
The View-Epipolar Interaction Module (VEI) is designed as a dual-branch structure, with one branch comprising the View Transformer and the other the Along-Epipolar Perception. The VEI input $\boldsymbol{X} \in \mathbb{R}^{N \times M \times C}$ comes from the CNN feature map interpolated features or from the output of the previous layer of EVI, and the VEI output $\boldsymbol{Y_{VEI}}$ is used as the input to the current layer of EVI.

\noindent\textbf{View Transformer.}\; The View Transformer is responsible for aggregating features across view dimensions. The view transformer takes the input $\boldsymbol{X}$, allowing it to perform self-attention operations in the view dimension ($M$).  To be more specific, the query, key, and value matrices are computed using linear mappings:
\begin{gather}
\boldsymbol{Q}=\boldsymbol{X}\boldsymbol{W_{Q}},\,\boldsymbol{K}=\boldsymbol{X}\boldsymbol{W_{K}},\,\boldsymbol{V}=\boldsymbol{X}\boldsymbol{W_{V}},
\end{gather}
where $\boldsymbol{W_{Q}},\boldsymbol{W_{K}},\boldsymbol{W_{V}}\in \mathbb{R}^{C\times C}$ are the linear mappings without biases. These matrices are then split into $h$ heads $\boldsymbol{Q}=[\boldsymbol{Q}^{1},\cdots,\boldsymbol{Q}^{h}]$, $\boldsymbol{K}=[\boldsymbol{K}^{1},\cdots,\boldsymbol{K}^{h}]$, and $\boldsymbol{V}=[\boldsymbol{V}^{1},\cdots,\boldsymbol{V}^{h}]$, each with $d=C/h$ channels. To enable the model to learn the relative spatial relationships between the target view and the source views,  we integrate the differences $\Delta \boldsymbol{d}^{s}$ (see Appendix \ref{Difference of Views}) between the target view and the source views into the self-attention mechanism:
\begin{gather}
    \boldsymbol{\tilde{X}}^{i} = \text{softmax}\left(\left(\boldsymbol{Q}^{i}+\Delta \boldsymbol{d}^{s}\right)\left(\boldsymbol{K^{i}}\right)^\top\right) \boldsymbol{V}^{i}.
\end{gather}

Ultimately, we obtain $\boldsymbol{\tilde{X}}=[\boldsymbol{\tilde{X}}^{i},\cdots,\boldsymbol{\tilde{X}}^{h}]$, and we employ a conventional Feed-Forward Network (FFN) to perform point-wise feature transformation:
\begin{gather}
    \boldsymbol{Y} = \text{FFN}(\boldsymbol{\tilde{X}})+\boldsymbol{\tilde{X}}.
\end{gather}

\noindent\textbf{Along-Epipolar Perception.}\; The Along-Epipolar Perception, serving as the second branch of VEI, aims to extract view-independent depth information to provide appearance continuity prior to the 3D representation. We compute the mean and variance of $\boldsymbol{V}\in \mathbb{R}^{N\times M\times C}$ in the view dimension ($M$) within the view transformer to obtain the global view-independent  feature $\boldsymbol{f}^{0}\in \mathbb{R}^{N\times 2C}$. We proceed to perceive the depth information along the entire ray through an adjacent-depth attention (1D Convolution AE) in the ray dimension ($N$). Since the information along an epipolar line is inherently continuous, a convolution operation that is seen as a kind of adjacent attention can learn the appearance continuity prior, which predicts the importance weights $\operatorname{w}_{i}^{v}$ for the sampling points:
\begin{gather}
    \boldsymbol{f}^{1} = \text{concat}\left(\boldsymbol{f}^{0}, \boldsymbol{x}, \boldsymbol{d}\right),\notag\\
    \{\operatorname{w}_{i}^{v}\}_{i=1}^{N} = \text{sigmoid}\left(\text{AE}\left(\{\boldsymbol{f}^{1}_{i}\}_{i=1}^{N}\right)\right), 
\end{gather}
where $\boldsymbol{x}$ and $\boldsymbol{d}$ refer to the 3D point position and the direction of the target ray’s sampling point. Particularly, $\boldsymbol{d}$ is copied to the same dimension as $\boldsymbol{x}$. GeoNeRF~\cite{johari2022geonerf} also employs an AE network to predict  coherent volume densities. However, our approach is more similar to an adjacent attention mechanism predicting depth importance weights and learning appearance continuity prior based on global epipolar features.

Combining the output of the View Transformer and Along-Epipolar Perception, the final output of VEI is calculated as follows:
\begin{gather}
    \boldsymbol{Y_{VEI}} = \boldsymbol{w}^{v} \cdot \boldsymbol{Y},
\end{gather}
where $\boldsymbol{w}^{v}=[\operatorname{w}_{1}^{v},\cdots,\operatorname{w}_{N}^{v}]$,  $\boldsymbol{Y_{VEI}}$ denotes the VEI's output, and $\cdot$ denotes element-wise multiplication.

\subsection{Epipolar-View Interaction}
Similar to VEI, The Epipolar-View Interaction Module (EVI) consists of two branches, the Epipolar Transformer and the Multi-View Calibration. The EVI input $\boldsymbol{X'} \in \mathbb{R}^{M \times N \times C}$ comes from the output of the current layer of VEI, and the EVI output $\boldsymbol{Y_{EVI}}$ is used as the input to the next layer of EVIs or as the total output of the aggregation network.\vspace{8pt}

\noindent\textbf{Epipolar Transformer.}\; The Epipolar Transformer takes the input $\boldsymbol{X'}$, enabling self-attention operations in the epipolar dimension ($N$). In particular, the epipolar transformer shares the same network structure as the view transformer above:
\begin{gather}
\boldsymbol{Q'}=\boldsymbol{X'}\boldsymbol{W'_{Q}},\,\boldsymbol{K'}=\boldsymbol{X'}\boldsymbol{W'_{K}},\,\boldsymbol{V'}=\boldsymbol{X'}\boldsymbol{W'_{V}}, \notag\\
\boldsymbol{\tilde{X'}}^{i} = \text{softmax}\left(\left(\boldsymbol{Q'}^{i}+\Delta \boldsymbol{d'}^{s}\right)\left(\boldsymbol{K'}^{i}\right)^\top\right) \boldsymbol{V'}^{i}, \notag\\
\boldsymbol{Y'} = \text{FFN}(\boldsymbol{\tilde{X'}})+\boldsymbol{\tilde{X'}}, 
\end{gather}
where $\boldsymbol{X}'[i ,j, k] = \boldsymbol{X}[j ,i, k]$, $\boldsymbol{d'}^{s}[i ,j, k] = \boldsymbol{d}^{s}[j ,i, k]$, $i, j, k$ denote the 1st ($M$), 2nd ($N$), and 3rd dimensions ($C$) respectively. \vspace{8pt}

\noindent\textbf{Multi-View Calibration.}\; The Multi-View Calibration, serving as the second branch of the EVI module, is employed to aggregate cross-view features and provide geometry consistency prior, aiming at calibrating the epipolar features. We calculate the weight values $\operatorname{w}_{j}^{e}$ for the target rays in each source view using the cross-view attention mechanism. In this process, we utilize $\boldsymbol{V}'\in \mathbb{R}^{M\times N\times C}$ from the epipolar transformer as the input:
\begin{gather}
    \boldsymbol{q}=\text{max}(\boldsymbol{V}') + \text{linear}(\Delta\boldsymbol{pose}),\notag \\
    \{\operatorname{w}_{j}^{e}\}_{j=1}^{M} = \text{sigmoid}\left(\text{Self-Attn}\left(\boldsymbol{q}, \boldsymbol{q}, \boldsymbol{q}\right)\right), 
\end{gather}
where $\Delta\boldsymbol{pose}$ (see Appendix \ref{Difference of Camera Poses}) refers to the difference between the source view camera pose and the target view camera pose, and $\text{linear}$ denotes the linear layer. Ultimately, incorporating the regression results of multi-view calibration, the output of the EVI is calculated as follows:
\begin{gather}
    \boldsymbol{Y_{EVI}} = \boldsymbol{w}^{e} \cdot  \boldsymbol{Y'},
\end{gather}
where $\boldsymbol{w}^{e}=[\operatorname{w}_{1}^{e},\cdots,\operatorname{w}_{M}^{e}]$, $\boldsymbol{Y_{EVI}}$ denotes the EVI's output, and $\cdot$ denotes element-wise multiplication.

 \begin{figure}[t]
    \centering
    \begin{subfigure}{0.3\linewidth}
		\centering
		\includegraphics[width=\linewidth]{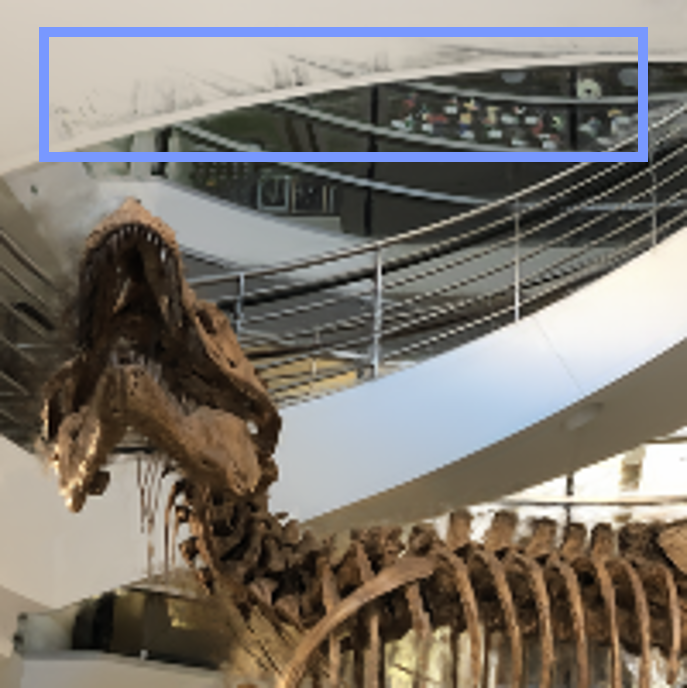}
		\caption{\centering only view transformer}
	\end{subfigure}
\centering
 \begin{subfigure}{0.3\linewidth}
		\centering
		\includegraphics[width=\linewidth]{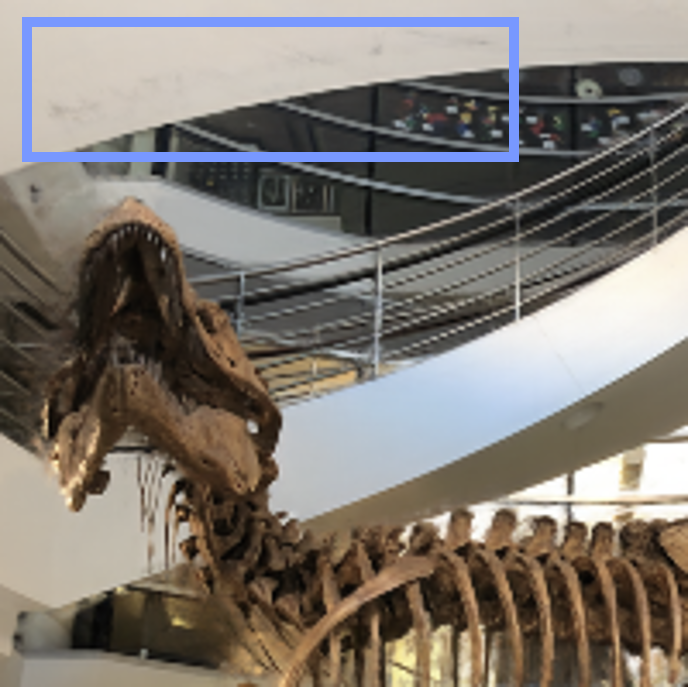}
		\caption{\centering w/ along-epipolar perception}
	\end{subfigure}
 \centering
 \begin{subfigure}{0.3\linewidth}
		\centering
		\includegraphics[width=\linewidth]{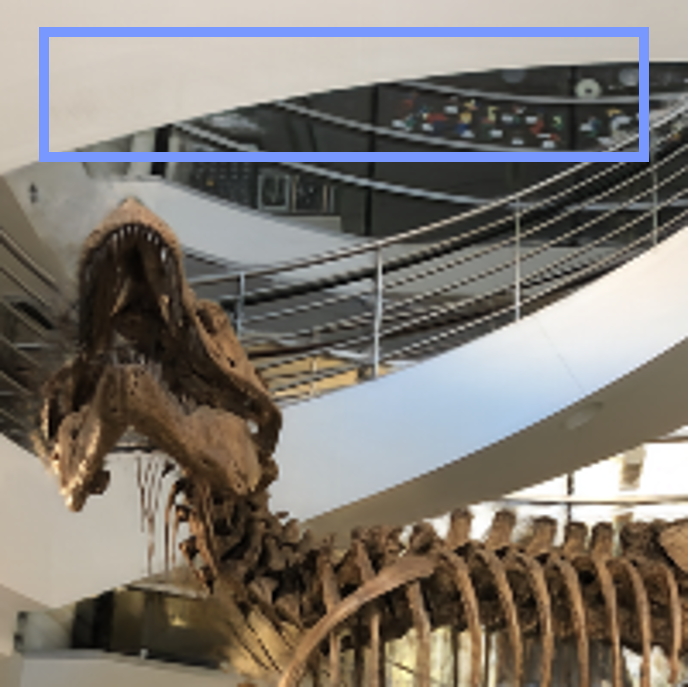}
		\caption{\centering EVE-NeRF\newline(ours)}
	\end{subfigure}
 
 \centering
    \begin{subfigure}{0.3\linewidth}
		\centering
		\includegraphics[width=\linewidth]{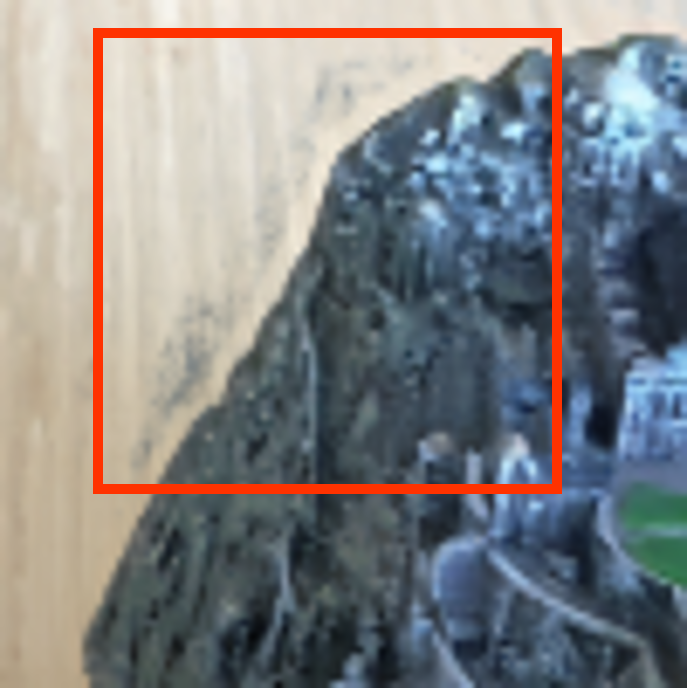}
		\caption{\centering only epipolar  transformer}
	\end{subfigure}
\centering
 \begin{subfigure}{0.3\linewidth}
		\centering
		\includegraphics[width=\linewidth]{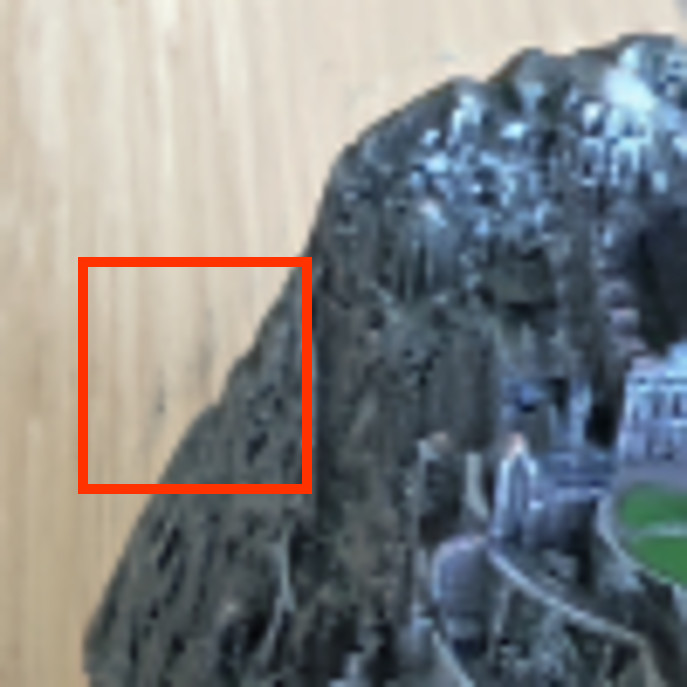}
		\caption{\centering w/ multi-view calibration}
	\end{subfigure}
 \centering
 \begin{subfigure}{0.3\linewidth}
		\centering
		\includegraphics[width=\linewidth]{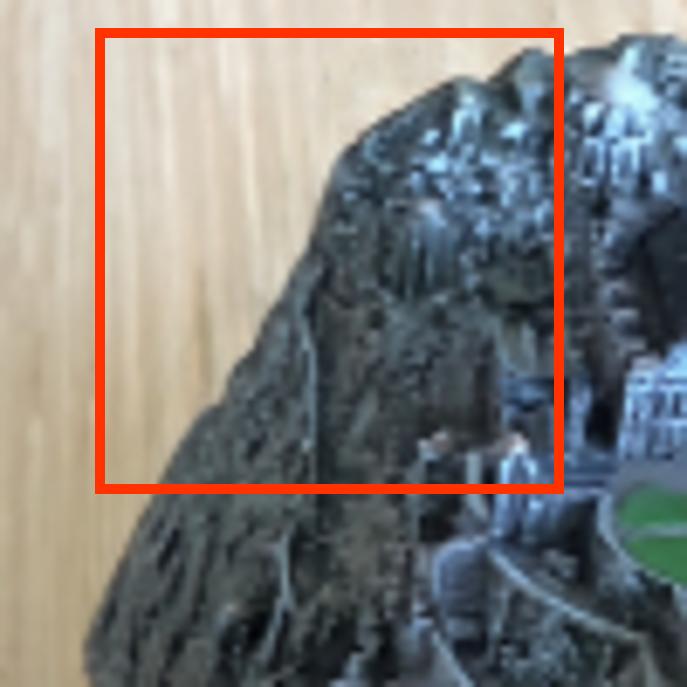}
		\caption{\centering EVE-NeRF\newline(ours)}
	\end{subfigure}
 \caption{
 The along-epipolar perception provides appearance continuity prior through adjacent-depth attention along the ray, while the multi-view calibration offers geometry consistency prior via cross-view attention. Our proposed method significantly reduces artifacts in rendering new views compared to single-dimension transformers.
 }
	\label{ablation_fig}
\end{figure}

\subsection{Conditioned NeRF Decoder}
We follow the established techniques of previous works~\cite{yu2021pixelnerf} to construct an MLP-based rendering network. We also condition 3D points on a ray using the generalizable 3D representation  $\boldsymbol{z}$ based on Eq. \ref{formula1}. Nevertheless, we diverge from the traditional MLP decoder~\cite{mildenhall2021nerf}, which processes each point on a ray independently. Instead, we take a more advanced approach by introducing cross-point interactions. For this purpose, we employ the ray Transformer from IBRNet~\cite{wang2021ibrnet} in our implementation.
After the rendering network predicts the emitted color $\boldsymbol{c}$ and volume density $\sigma$, we can generate target pixel color using volume rendering~\cite{mildenhall2021nerf}:
\begin{gather}
    \boldsymbol{C}=\sum_{i=1}^{N} T_{i}\left(1-\exp \left(-\sigma_{i} \delta_{i}\right)\right) \boldsymbol{c}_{i}, \quad T_{i}=\exp (-\sum_{j}^{i-1} \sigma_{j} \delta_{j}),
\end{gather}
where $\boldsymbol{c}_{i},\sigma_{i}$ which are calculated based on Eq. \ref{formula1}, refer to the color and density of the $i$-th sampling point on the ray. 

\subsection{Training Objectives}
EVE-NeRF is trained solely using a photometric loss function, without the need for additional ground-truth 3D data. Specifically, our training loss function is as follows:
\begin{gather}
    \mathcal{L}=\sum_{\boldsymbol{p} \in \mathcal{P}}\left\|\boldsymbol{C}_{pred}-\boldsymbol{C}_{gt}\right\|_{2}^{2},
\end{gather}
where $\mathcal{P}$ represents a set of pixel points in a training batch, $\boldsymbol{C}_{pred},\boldsymbol{C}_{gt}$ respectively represent the rendering color for pixel $\boldsymbol{p}$ and the ground-truth color.

\section{Experiments}
\label{sec:experiments}

\subsection{Implementation Details} 
We randomly sample $2,048$ rays per batch, each with $N=88$ sampling points along the rays. Our lightweight CNN and EVE-NeRF models are trained for $250,000$ iterations using an Adam optimizer with initial learning rates of $1e^{-3}$ and $5e^{-4}$, respectively, and an exponential learning rate decay. The training is performed end-to-end on 4 V100-32G GPUs for 3 days. 
To evaluate our model, we use common metrics such as PSNR, SSIM, and LPIPS and compare the results qualitatively and quantitatively with other generalizable neural rendering approaches. More details such as network hyperparameters are provided in Appendix \ref{Additional Technical Details}.

\subsection{Comparative Studies}
To provide a fair comparison with prior works~\cite{wang2021ibrnet,chen2021mvsnerf,varma2022attention}, we conducted experiments under 2 different settings: Generalizable NVS and Few-Shot NVS, as was done in GPNR~\cite{suhail2022generalizable}.\vspace{8pt}


\begin{figure*}[htbp]
\centering
    \begin{subfigure}{0.245\linewidth}
		\centering
		\includegraphics[width=\linewidth]{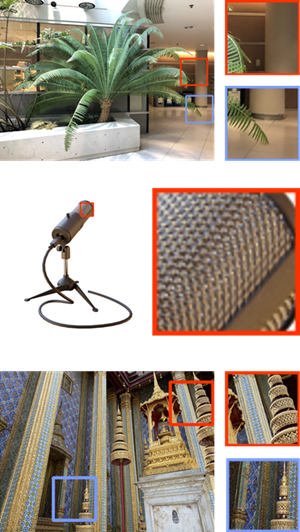}
		\caption{GT}
	\end{subfigure}
    \centering
    \begin{subfigure}{0.245\linewidth}
		\centering
		\includegraphics[width=\linewidth]{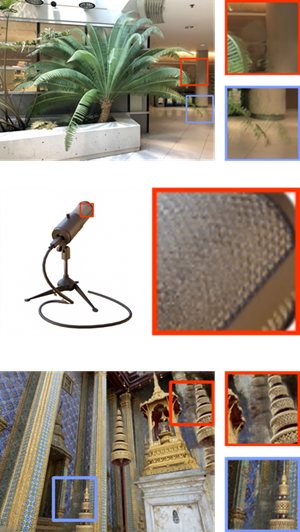}
		\caption{IBRNet}
	\end{subfigure}
\centering
 \begin{subfigure}{0.245\linewidth}
		\centering
		\includegraphics[width=\linewidth]{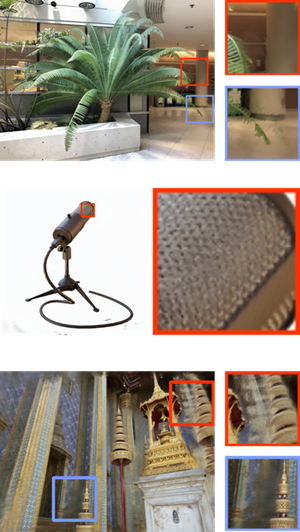}
		\caption{GNT}
	\end{subfigure}
 \centering
 \begin{subfigure}{0.245\linewidth}
		\centering
		\includegraphics[width=\linewidth]{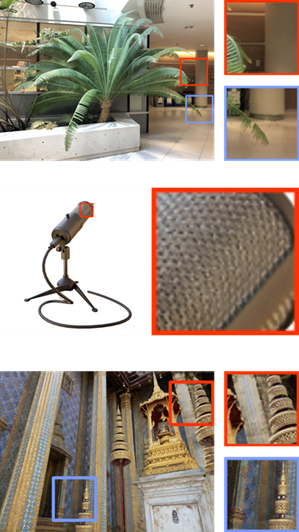}
		\caption{EVE-NeRF}
	\end{subfigure}
 \caption{Qualitative comparison of EVE-NeRF with IBRNet\cite{wang2021ibrnet} and GNT\cite{varma2022attention} in setting 1. The first, second, and third rows correspond to the Fern scene from LLFF, the Mic scene from Blender, and the Crest scene from Shiny, respectively. Our method, EVE-NeRF, demonstrates superior capability compared to the baselines in accurately reconstructing the geometry, appearance, and complex texture regions. In particular, our method successfully reconstructs the leaves and the surrounding area in the Fern scene.}
	\label{setting1_fig}
\end{figure*}
\begin{table*}
\centering

\begin{tabular}{l|ccc|ccc|ccc}
\toprule 
\multirow{2}{*}{Method} & & LLFF& &  &Blender & & &Shiny & \\  \cline{2-4} \cline{4-7}  \cline{8-10} 
& PSNR$\uparrow$ & SSIM$\uparrow$ & LPIPS$\downarrow$ & PSNR$\uparrow$ & SSIM$\uparrow$ & LPIPS$\downarrow$ & PSNR$\uparrow$ & SSIM$\uparrow$ & LPIPS$\downarrow$ \\ 
\midrule
PixelNeRF~\cite{yu2021pixelnerf} & 18.66 & 0.588 & 0.463  &22.65 &0.808 &0.202 &- &- &-\\
IBRNet~\cite{wang2021ibrnet} & 25.17 & 0.813 & 0.200  &26.73 &0.908 &0.101 &23.60  &0.785  &0.180 \\
GPNR~\cite{suhail2022generalizable} & 25.35 & 0.818 & 0.198  &\textbf{28.29} &0.927 &0.080 &25.72  &0.880  &0.175 \\
NeuRay~\cite{liu2022neural} & 25.72 & 0.880 & 0.175  &26.48 &0.944 &0.091 &24.12  &0.860  &0.170 \\
GNT~\cite{varma2022attention} & 25.53 & 0.836 & 0.178  &26.01 &0.925 &0.088 &27.19  &0.912  &0.083 \\
EVE-NeRF & \textbf{27.16} & \textbf{0.912} & \textbf{0.134}  &27.03 &\textbf{0.952} &\textbf{0.072} &\textbf{28.01}  &\textbf{0.935}  &\textbf{0.083}\\
\bottomrule
\end{tabular}

\caption{
Results for setting 1. Our proposed method, EVE-NeRF, outperforms most of the baselines on the majority of the metrics. With the exception of PixelNeRF~\cite{yu2021pixelnerf}, all baseline methods~\cite{wang2021ibrnet,suhail2022generalizable,liu2022neural} employ sequential or independent  transformer-based single-dimensional ray aggregation. In contrast, our approach is based on a dual-branch structure, enabling multi-dimensional interactions for both view and epipolar information. The results confirm that our method's multi-dimensional ray feature aggregation is superior to the single-dimensional aggregation used in the baselines.
}
\vspace{-0.05in}
\label{tab:results for setting 1}
\end{table*}


\noindent\textbf{Setting 1: Generalizable NVS.}\; Following IBRnet~\cite{wang2021ibrnet}, we set up the reference view pool comprising $k \times M$ proximate views.
$M$ views are chosen at random from this pool to serve as source views.  
Throughout the training process, the parameters $k$ and $M$ are subject to uniform random sampling, with $k$ drawn from  $(1, 3)$ and $M$ from  $(8, 12)$. During evaluation, we fix the number of source views $M=10$.

For the training dataset, we adopt object renderings of $1,030$ models from Google Scanned Object~\cite{downs2022google}, RealEstate10K~\cite{zhou2018stereo}, $100$ scenes from the Spaces dataset~\cite{flynn2019deepview} and $95$ real scenes from handheld cellphone captures~\cite{mildenhall2019local, wang2021ibrnet}. For the evaluation dataset, we use Real Forward-Facing~\cite{mildenhall2019local}, Blender~\cite{mildenhall2021nerf} and Shiny~\cite{wizadwongsa2021nex}. 

Table \ref{tab:results for setting 1} presents the quantitative results, while Figure \ref{setting1_fig} showcases the qualitative results. As shown in Table \ref{tab:results for setting 1}, comparison to methods~\cite{wang2021ibrnet,suhail2022generalizable,liu2022neural} using transformer networks for feature aggregation, our approach outperforms them under most metrics. Our method outperforms SOTA method GNT~\cite{varma2022attention} by $4.43\%$$\uparrow$ PSNR, $4.83\%$$\uparrow$ SSIM, $14.3\%$$\downarrow$ LPIPS in 3 evaluating dataset evenly. 
Such results verify the effectiveness of introducing the complementary appearance continuity and geometry consistency priors to the feature aggregation. 
Additionally, as shown in the second row of Figure \ref{setting1_fig}, our method successfully reconstructs intricate textural details.\vspace{8pt}

\noindent\textbf{Setting 2: Few-Shot NVS.}\; To compare with few-shot generalizable neural rendering methods~\cite{chen2021mvsnerf,chen2023explicit}, we conducted novel view synthesis experiments with $M=3$ input views in both training and evaluating, following the MVSNeRF~\cite{chen2021mvsnerf} setting. We split the DTU~\cite{jensen2014large} dataset into 88 scenes for training and 16 scenes for testing, following the methodology of prior works. Additionally, we also conducted training on the Google Scanned Object dataset.
As shown in Table \ref{tab:results for setting 2}, we performed a quantitative comparison with 4 few-shot generalizable neural rendering methods~\cite{yu2021pixelnerf,wang2021ibrnet,chen2021mvsnerf,chen2023explicit} on the DTU testing dataset and Blender. With only 3 source views input for setting, our model still achieves good performance. 
Our method outperforms SOTA methods MatchNeRF~\cite{chen2023explicit} by $2.19\%$$\uparrow$ PSNR,  $13.0\%$$\downarrow$ LPIPS in 2 evaluating dataset evenly. Please refer to Appendix \ref{Qualitative Comparison for Setting 2} for the qualitative comparison for setting 2.

\noindent\textbf{Efficiency Comparison.}\; As shown in the Tab \ref{tab:fps_compare}, our method is compared with GNT~\cite{varma2022attention} and GPNR~\cite{suhail2022generalizable} in terms of efficiency. We perform the testing of setting 1 in  LLFF dataset. The result illustrates that our method not only requires less memory and faster rendering time per-image, but also has a higher PSNR for novel view synthesis.

\begin{figure*}
\begin{minipage}{\textwidth}

\begin{minipage}[t]{0.64\textwidth}
\makeatletter\def\@captype{table}
\centering
\begin{tabular}{l|lll|lll}
\toprule 
\small\multirow{2}{*}{Method} & & \small DTU& &  &\small Blender & \\  \cline{2-4} \cline{4-7} 
& \small PSNR$\uparrow$ & \small SSIM$\uparrow$ & \small LPIPS$\downarrow$ & \small PSNR$\uparrow$ & \small SSIM$\uparrow$ & \small LPIPS$\downarrow$ \\
\midrule
\small PixelNeRF~\cite{yu2021pixelnerf} & \small19.31 & \small0.789 & \small0.671  &\small7.39 &\small0.658 &\small0.411 \\
\small IBRNet~\cite{wang2021ibrnet} & \small26.04 & \small0.917 & \small0.190  &\small22.44 &\small0.874 &\small0.195 \\
\small MVSNeRF~\cite{chen2021mvsnerf} & \small26.63 & \small0.931 & \small0.168  &\small\textbf{23.62} &\small0.897 &\small0.176 \\
\small  MatchNeRF~\cite{chen2023explicit} & \small26.91 & \small0.934 & \small0.159  &\small23.20 &\small0.897 &\small0.164 \\
\small EVE-NeRF & \small\textbf{27.80} & \small\textbf{0.937} & \small\textbf{0.149}  &\small23.45 &\small\textbf{0.903} &\small\textbf{0.132} \\
\bottomrule
\end{tabular}
\captionsetup{margin=1em}
\caption{
Results for setting 2.  Our method (EVE-NeRF) is trained on DTU and the Google Scanned Object dataset with 3 reference views. Our method outperforms on multiple metrics with other few-shot generalizable neural rendering methods.
}
\vspace{-0.05in}
\label{tab:results for setting 2}
\end{minipage}
\begin{minipage}[t]{0.34\textwidth}
\makeatletter\def\@captype{table}
\centering
\vspace{-50pt}
    \begin{tabular}{lccc}
\toprule 
 \small Model &  \small Storage$\downarrow$ & \small Time$\downarrow$   & \small PSNR$\uparrow$ \\           
\midrule
\small GNT~\cite{varma2022attention} & \small 112MB & \small 208s & \small 25.35 \\
\small GPNR~\cite{suhail2022generalizable} & \small 110MB & \small 12min & \small 25.53 \\
 \small EVE-NeRF & \small \textbf{53.8MB} & \small \textbf{194s}& \small \textbf{27.16} \\
\bottomrule
\end{tabular}

\captionsetup{margin=0em}
\caption{
Efficiency comparison results in LLFF dataset on the same RTX4090. Our method requires less storage, shorter rendering time per new view synthesis, and higher quality reconstruction compared to GNT~\cite{varma2022attention} and GPNR~\cite{suhail2022generalizable}.
}
\vspace{-0.05in}
\label{tab:fps_compare}

\end{minipage}
\end{minipage}
\end{figure*}




\subsection{Ablation Studies}
\label{ablation}
To evaluate the significance of our contributions, we conducted extensive ablation experiments. We trained on setting 1 and tested on the Real Forward-Facing dataset. For efficiency, in both the training and testing datasets we set the resolutions of images reduced by half in both the training and testing datasets, resulting in the resolution of $504 \times 378$ for the Real Forward-Facing dataset.\vspace{8pt}

\noindent \textbf{Only view/epipolar transformer.}\; 
In this ablation, we maintain only the view/epipolar transformers and NeRF decoder.
As can be observed in the first row of Table \ref{tab:results for ablation},  using only view/epipolar transformer reduces PSNR by  $6.21\%$ compared to EVE-NeRF due to the limitations of view/epipolar aggregation in only one dimension.
\vspace{8pt}

\noindent \textbf{Along-epipolar perception.}\; 
Compared with only view transformer, we retained view transformer with along-epipolar perception in this ablation. As shown in Table \ref{tab:results for ablation} and Figure \ref{ablation_fig}, using along-epipolar perception would increase PSNR by $1.80\%$. The appearance continuity prior provided by along-epipolar perception compensates for the missing epipolar information in the pure view aggregation model.\vspace{8pt}

\noindent \textbf{Multi-view calibration.}\; 
Similarly, against to only epipolar transformer, we kept the epipolar transformer with multi-view calibration. As can be observed in Table \ref{tab:results for ablation} and Figure \ref{ablation_fig}, adopting multi-view calibration would improve the performance of generalizable rendering. 
It verifies that the multi-view calibration can enhance epipolar aggregating ability via geometry prior.\vspace{8pt}

\noindent \textbf{Na\"ive dual network architecture.}\; 
To validate the effectiveness of our proposed entangled aggregation, we compared our EVE-NeRF with two other architectures: the Na\"ive Dual Transformer and the Dual Transformer with Cross-Attention Interaction  (see Appendix \ref{Additional Technical Details}). 
Our method outperforms both na\"ive dual transformer by $3.85\%$  and $3.14\%$ PSNR. Visualization results are provided in Appendix \ref{Qualitative Comparison With naive}. It is demonstrated that our proposed EVE-NeRF has a robust ability to fusing view and epipolar pattern.

\begin{table}[t]
\centering
\resizebox{\columnwidth}{!}{

\begin{tabular}{lccc}
\toprule 
Model & PSNR$\uparrow$ & SSIM$\uparrow$ & LPIPS$\downarrow$ \\           
\midrule
only view transformer & 25.03 & 0.886 & 0.132 \\
+ along-epipolar perception & 25.48 & 0.892 & 0.128 \\
only epipolar transformer & 25.02 & 0.879 & 0.147 \\
+ multi-view calibration & 25.17 & 0.883 & 0.141 \\
naïve dual transformer & 25.66 & 0.890 & 0.128 \\
+ cross-attention interaction & 25.85 & 0.896 & 0.120 \\
\midrule
EVE-NeRF & \textbf{26.69} & \textbf{0.913} & \textbf{0.102} \\
\bottomrule
\end{tabular}

}
\caption{
Ablations. The ablation study was conducted by training in a low-resolution setting 1 and testing on LLFF dataset with the resolution of $504 \times 378$. 
}
\vspace{-0.05in}
\label{tab:results for ablation}
\end{table}




\subsection{Visualization on Entangled Information Interaction}
To further validate the entangled information interaction module's ability of providing the \emph{de facto} appearance continuity prior and geometry consistency prior, we visualize and analyze the importance weights predicted by the along-epipolar perception and multi-view calibration.

\begin{figure}[t]
    \centering
    \begin{subfigure}{0.49\linewidth}
		\centering
		\includegraphics[width=\linewidth]{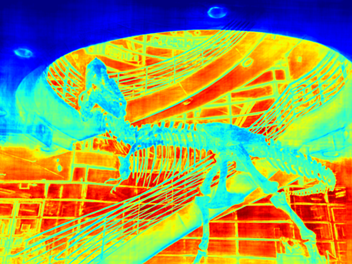}
		\caption{Adjacent-depth attention map}
		\label{depth_attn2}
	\end{subfigure}
\centering
 \begin{subfigure}{0.49\linewidth}
		\centering
		\includegraphics[width=\linewidth]{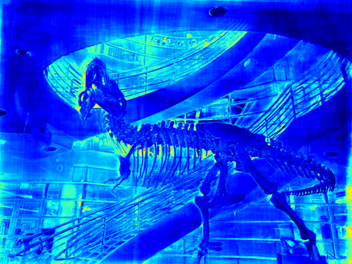}
		\caption{Rendered depth map}
		\label{depth_attn3}
	\end{subfigure}
 \caption{Visualizations of adjacent-depth attention map and rendered depth map. By capitalizing on the appearance continuity prior, adjacent-depth attention boosts the coherence in depth map.}
	\label{depth_attn}
\end{figure}

The along-epipolar perception provides appearance continuity prior and regresses the importance weights for the target ray's sampled depths. Specifically, we obtain a depth map by multiplying the depth weights with the marching distance along the ray. 
As shown in Figure \ref{depth_attn}, the adjacent-depth attention map demonstrates a more coherent character, indicating that the along-epipolar perception provides beneficial appearance consistency prior.

 \begin{figure}[t]
    \centering
    \begin{subfigure}{0.49\linewidth}
		\centering
		\includegraphics[width=\linewidth]{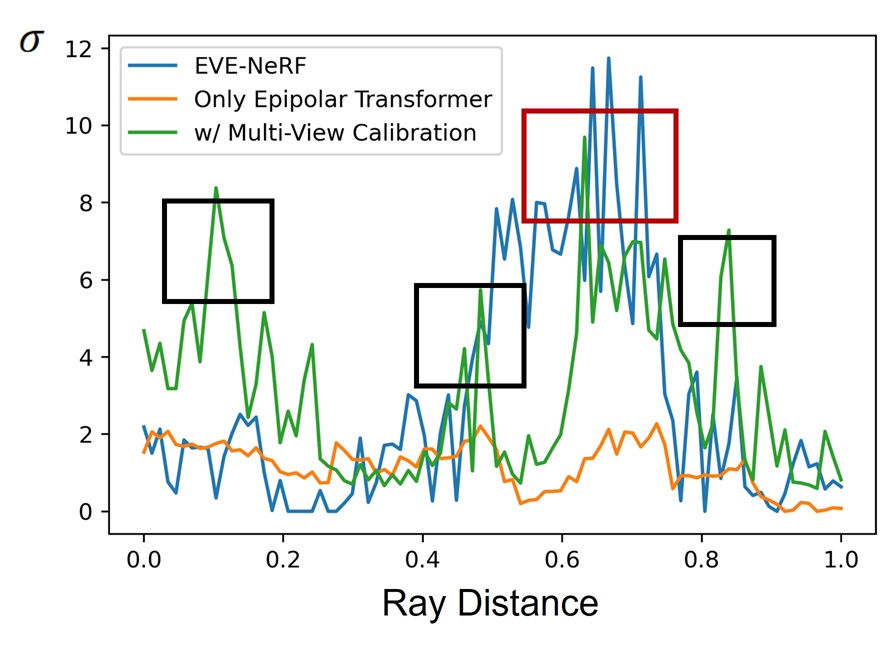}
		\caption{Volume density of ray 1}
		\label{plot_ray1}
	\end{subfigure}
\centering
 \begin{subfigure}{0.49\linewidth}
		\centering
		\includegraphics[width=\linewidth]{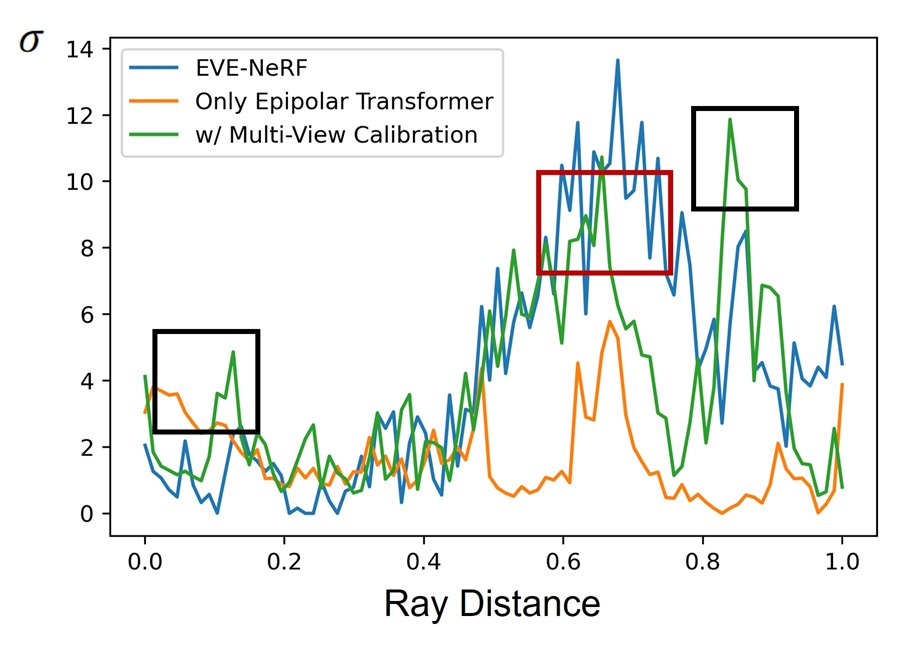}
		\caption{Volume density of ray 2}
		\label{plot_ray2}
	\end{subfigure}
 \caption{Line charts of the volume density from a novel view of the Chair~\cite{mildenhall2021nerf}. Red boxes represent correct peaks and black boxes represent abnormal peaks. Multi-View Calibration learns more complex signals but with more noise. Volume densities predicted by EVE-NeRF are more likely to have a single-peak distribution.}
	\label{plot_ray}
\end{figure}

\begin{figure}[t]
    \centering
    \begin{subfigure}{0.412\linewidth}
		\centering
		\includegraphics[width=\linewidth]{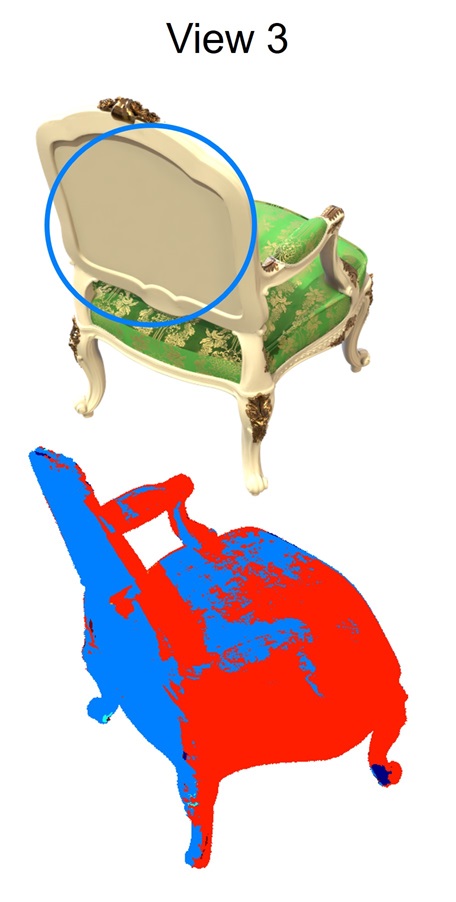}
		\caption{View Transformer}
		\label{view_attn1}
	\end{subfigure}
\centering
 \begin{subfigure}{0.412\linewidth}
		\centering
		\includegraphics[width=\linewidth]{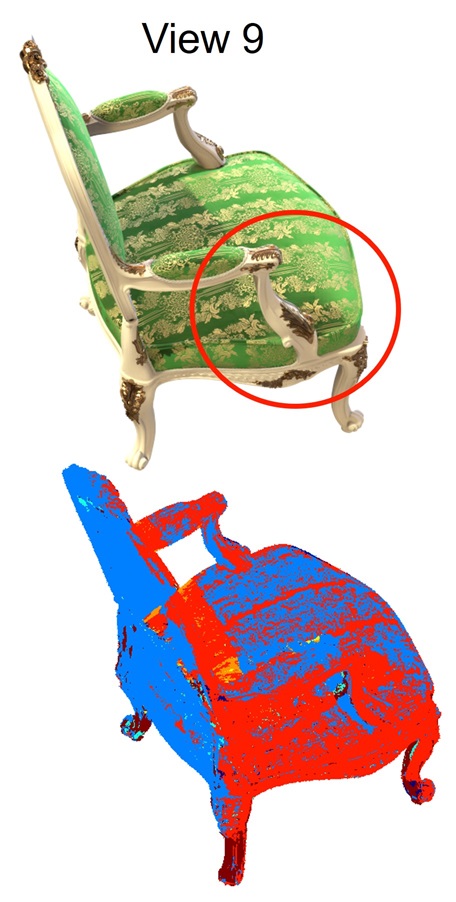}
		\caption{Multi-View Calibration}
		\label{view_attn2}
	\end{subfigure}
 \centering
 \begin{subfigure}{0.146\linewidth}
		\centering
		\includegraphics[width=\linewidth]{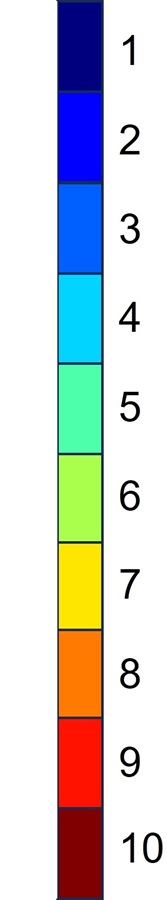}
		\caption{}
		\label{view_attn3}
	\end{subfigure}
 \caption{Each color represents the source view ID corresponding to the maximum weight for the target pixel. 
 Both the view transformer and the multi-view calibration have successfully learned the cross-view information from the source views.}
	\label{view_attn}
\end{figure}

The multi-view calibration provides geometry consistency prior and predicts the importance weights for the source views. 
As shown in Figure \ref{plot_ray}, we visualize two line charts of the point density. Multi-view calibration learns more complex light signals, but with a multi-peaked distribution. EVE-NeRF predicts point density distributions with distinct peaks and reduced noise in the light signals.
As shown in Figure \ref{view_attn}, 
the view transformer  and the multi-view calibration correctly predict the correspondence between the target pixel and the source views, such as the back of the chair. Furthermore, both methods predict that the pixels in the upper right part of the chair correspond to source view 3, where the upper right part of the chair is occluded. We believe that EVE-NeRF learns about the awareness of visibility, even when the target pixel is occluded. 




\section{Conclusion}
\label{sec:conclusion}
We propose a new Generalizable NeRF named EVE-NeRF that aggregates cross-view and along-epipolar information in an entangled manner. The core of EVE-NeRF consists of our new proposed View Epipolar Interaction Module (VEI) and Epipolar View Interaction Module (EVI) that are organized alternately. VEI and EVI can project the scene-invariant appearance continuity and geometry consistency priors, which serve to offset information losses that typically arises from single-dimensional aggregation operations. We demonstrate the superiority of our method in both generalizable and few-shot NVS settings compared with the state-of-the-art methods. Additionally, extensive ablation studies confirm that VEI and EVI can enhance information interaction across view and epipolar dimensions to yield better generalizable 3D representation.


\section*{Acknowledgement}
\label{sec:acknowledment}
This work is supported by the National Natural Science Foundation of China (62293554, 62206249, U2336212), the Natural Science Foundation of Zhejiang Province, China (LZ24F020002), Young Elite Scientists Sponsorship Program by CAST (2023QNRC001).

{\small
\bibliographystyle{ieeenat_fullname}
\bibliography{11_references}

\begin{thebibliography}{66}
\providecommand{\natexlab}[1]{#1}
\providecommand{\url}[1]{\texttt{#1}}
\expandafter\ifx\csname urlstyle\endcsname\relax
  \providecommand{\doi}[1]{doi: #1}\else
  \providecommand{\doi}{doi: \begingroup \urlstyle{rm}\Url}\fi

\bibitem[Barron et~al.(2021)Barron, Mildenhall, Tancik, Hedman, Martin-Brualla, and Srinivasan]{barron2021mip}
Jonathan~T Barron, Ben Mildenhall, Matthew Tancik, Peter Hedman, Ricardo Martin-Brualla, and Pratul~P Srinivasan.
\newblock Mip-nerf: A multiscale representation for anti-aliasing neural radiance fields.
\newblock In \emph{Proceedings of the IEEE/CVF International Conference on Computer Vision}, pages 5855--5864, 2021.

\bibitem[Barron et~al.(2022)Barron, Mildenhall, Verbin, Srinivasan, and Hedman]{barron2022mip}
Jonathan~T Barron, Ben Mildenhall, Dor Verbin, Pratul~P Srinivasan, and Peter Hedman.
\newblock Mip-nerf 360: Unbounded anti-aliased neural radiance fields.
\newblock In \emph{Proceedings of the IEEE/CVF Conference on Computer Vision and Pattern Recognition}, pages 5470--5479, 2022.

\bibitem[Barron et~al.(2023)Barron, Mildenhall, Verbin, Srinivasan, and Hedman]{barron2023zip}
Jonathan~T Barron, Ben Mildenhall, Dor Verbin, Pratul~P Srinivasan, and Peter Hedman.
\newblock Zip-nerf: Anti-aliased grid-based neural radiance fields.
\newblock \emph{arXiv preprint arXiv:2304.06706}, 2023.

\bibitem[Chen et~al.(2021{\natexlab{a}})Chen, Xu, Zhao, Zhang, Xiang, Yu, and Su]{chen2021mvsnerf}
Anpei Chen, Zexiang Xu, Fuqiang Zhao, Xiaoshuai Zhang, Fanbo Xiang, Jingyi Yu, and Hao Su.
\newblock Mvsnerf: Fast generalizable radiance field reconstruction from multi-view stereo.
\newblock In \emph{Proceedings of the IEEE/CVF International Conference on Computer Vision}, pages 14124--14133, 2021{\natexlab{a}}.

\bibitem[Chen et~al.(2022)Chen, Xu, Geiger, Yu, and Su]{chen2022tensorf}
Anpei Chen, Zexiang Xu, Andreas Geiger, Jingyi Yu, and Hao Su.
\newblock Tensorf: Tensorial radiance fields.
\newblock In \emph{European Conference on Computer Vision}, pages 333--350. Springer, 2022.

\bibitem[Chen et~al.(2021{\natexlab{b}})Chen, Fan, and Panda]{chen2021crossvit}
Chun-Fu~Richard Chen, Quanfu Fan, and Rameswar Panda.
\newblock Crossvit: Cross-attention multi-scale vision transformer for image classification.
\newblock In \emph{Proceedings of the IEEE/CVF international conference on computer vision}, pages 357--366, 2021{\natexlab{b}}.

\bibitem[Chen et~al.(2023{\natexlab{a}})Chen, Xu, Wu, Zheng, Cham, and Cai]{chen2023explicit}
Yuedong Chen, Haofei Xu, Qianyi Wu, Chuanxia Zheng, Tat-Jen Cham, and Jianfei Cai.
\newblock Explicit correspondence matching for generalizable neural radiance fields.
\newblock \emph{arXiv preprint arXiv:2304.12294}, 2023{\natexlab{a}}.

\bibitem[Chen et~al.(2023{\natexlab{b}})Chen, Funkhouser, Hedman, and Tagliasacchi]{chen2023mobilenerf}
Zhiqin Chen, Thomas Funkhouser, Peter Hedman, and Andrea Tagliasacchi.
\newblock Mobilenerf: Exploiting the polygon rasterization pipeline for efficient neural field rendering on mobile architectures.
\newblock In \emph{Proceedings of the IEEE/CVF Conference on Computer Vision and Pattern Recognition}, pages 16569--16578, 2023{\natexlab{b}}.

\bibitem[Chen et~al.(2023{\natexlab{c}})Chen, Zhang, Gu, Kong, Yang, and Yu]{chen2023dual}
Zheng Chen, Yulun Zhang, Jinjin Gu, Linghe Kong, Xiaokang Yang, and Fisher Yu.
\newblock Dual aggregation transformer for image super-resolution.
\newblock In \emph{Proceedings of the IEEE/CVF International Conference on Computer Vision}, pages 12312--12321, 2023{\natexlab{c}}.

\bibitem[Chibane et~al.(2021)Chibane, Bansal, Lazova, and Pons-Moll]{chibane2021stereo}
Julian Chibane, Aayush Bansal, Verica Lazova, and Gerard Pons-Moll.
\newblock Stereo radiance fields (srf): Learning view synthesis for sparse views of novel scenes.
\newblock In \emph{Proceedings of the IEEE/CVF Conference on Computer Vision and Pattern Recognition}, pages 7911--7920, 2021.

\bibitem[Cong et~al.(2023)Cong, Liang, Wang, Fan, Chen, Varma, Wang, and Wang]{cong2023enhancing}
Wenyan Cong, Hanxue Liang, Peihao Wang, Zhiwen Fan, Tianlong Chen, Mukund Varma, Yi Wang, and Zhangyang Wang.
\newblock Enhancing nerf akin to enhancing llms: Generalizable nerf transformer with mixture-of-view-experts.
\newblock In \emph{Proceedings of the IEEE/CVF International Conference on Computer Vision}, pages 3193--3204, 2023.

\bibitem[Downs et~al.(2022)Downs, Francis, Koenig, Kinman, Hickman, Reymann, McHugh, and Vanhoucke]{downs2022google}
Laura Downs, Anthony Francis, Nate Koenig, Brandon Kinman, Ryan Hickman, Krista Reymann, Thomas~B McHugh, and Vincent Vanhoucke.
\newblock Google scanned objects: A high-quality dataset of 3d scanned household items.
\newblock In \emph{2022 International Conference on Robotics and Automation (ICRA)}, pages 2553--2560. IEEE, 2022.

\bibitem[Feng et~al.(2023)Feng, Wang, Wang, Yang, and Zheng]{feng2023clustering}
Tuo Feng, Wenguan Wang, Xiaohan Wang, Yi Yang, and Qinghua Zheng.
\newblock Clustering based point cloud representation learning for 3d analysis.
\newblock In \emph{CVPR}, pages 8283--8294, 2023.

\bibitem[Flynn et~al.(2019)Flynn, Broxton, Debevec, DuVall, Fyffe, Overbeck, Snavely, and Tucker]{flynn2019deepview}
John Flynn, Michael Broxton, Paul Debevec, Matthew DuVall, Graham Fyffe, Ryan Overbeck, Noah Snavely, and Richard Tucker.
\newblock Deepview: View synthesis with learned gradient descent.
\newblock In \emph{Proceedings of the IEEE/CVF Conference on Computer Vision and Pattern Recognition}, pages 2367--2376, 2019.

\bibitem[Gao et~al.(2021)Gao, Saraf, Kopf, and Huang]{gao2021dynamic}
Chen Gao, Ayush Saraf, Johannes Kopf, and Jia-Bin Huang.
\newblock Dynamic view synthesis from dynamic monocular video.
\newblock In \emph{Proceedings of the IEEE/CVF International Conference on Computer Vision}, pages 5712--5721, 2021.

\bibitem[Guan et~al.(2021)Guan, Huang, Luo, Liu, Xu, and Yang]{guan2021discriminative}
Qingji Guan, Yaping Huang, Yawei Luo, Ping Liu, Mingliang Xu, and Yi Yang.
\newblock Discriminative feature learning for thorax disease classification in chest x-ray images.
\newblock \emph{IEEE Transactions on Image Processing}, 30:\penalty0 2476--2487, 2021.

\bibitem[Hedman et~al.(2021)Hedman, Srinivasan, Mildenhall, Barron, and Debevec]{hedman2021baking}
Peter Hedman, Pratul~P Srinivasan, Ben Mildenhall, Jonathan~T Barron, and Paul Debevec.
\newblock Baking neural radiance fields for real-time view synthesis.
\newblock In \emph{Proceedings of the IEEE/CVF International Conference on Computer Vision}, pages 5875--5884, 2021.

\bibitem[Hu et~al.(2023)Hu, Wang, Ma, Yang, Gao, Liu, and Ma]{hu2023tri}
Wenbo Hu, Yuling Wang, Lin Ma, Bangbang Yang, Lin Gao, Xiao Liu, and Yuewen Ma.
\newblock Tri-miprf: Tri-mip representation for efficient anti-aliasing neural radiance fields.
\newblock In \emph{Proceedings of the IEEE/CVF International Conference on Computer Vision}, pages 19774--19783, 2023.

\bibitem[Huang et~al.(2023)Huang, Zhang, Feng, Li, Wang, and Wang]{huang2023local}
Xin Huang, Qi Zhang, Ying Feng, Xiaoyu Li, Xuan Wang, and Qing Wang.
\newblock Local implicit ray function for generalizable radiance field representation.
\newblock In \emph{Proceedings of the IEEE/CVF Conference on Computer Vision and Pattern Recognition}, pages 97--107, 2023.

\bibitem[Jensen et~al.(2014)Jensen, Dahl, Vogiatzis, Tola, and Aan{\ae}s]{jensen2014large}
Rasmus Jensen, Anders Dahl, George Vogiatzis, Engin Tola, and Henrik Aan{\ae}s.
\newblock Large scale multi-view stereopsis evaluation.
\newblock In \emph{Proceedings of the IEEE conference on computer vision and pattern recognition}, pages 406--413, 2014.

\bibitem[Johari et~al.(2022)Johari, Lepoittevin, and Fleuret]{johari2022geonerf}
Mohammad~Mahdi Johari, Yann Lepoittevin, and Fran{\c{c}}ois Fleuret.
\newblock Geonerf: Generalizing nerf with geometry priors.
\newblock In \emph{Proceedings of the IEEE/CVF Conference on Computer Vision and Pattern Recognition}, pages 18365--18375, 2022.

\bibitem[Kulh{\'a}nek et~al.(2022)Kulh{\'a}nek, Derner, Sattler, and Babu{\v{s}}ka]{kulhanek2022viewformer}
Jon{\'a}{\v{s}} Kulh{\'a}nek, Erik Derner, Torsten Sattler, and Robert Babu{\v{s}}ka.
\newblock Viewformer: Nerf-free neural rendering from few images using transformers.
\newblock In \emph{European Conference on Computer Vision}, pages 198--216. Springer, 2022.

\bibitem[Li et~al.(2021)Li, Feng, She, Ding, Wang, and Lee]{li2021mine}
Jiaxin Li, Zijian Feng, Qi She, Henghui Ding, Changhu Wang, and Gim~Hee Lee.
\newblock Mine: Towards continuous depth mpi with nerf for novel view synthesis.
\newblock In \emph{Proceedings of the IEEE/CVF International Conference on Computer Vision}, pages 12578--12588, 2021.

\bibitem[Liu et~al.(2021)Liu, Zhang, Zhang, Zhang, Zhu, and Russell]{liu2021editing}
Steven Liu, Xiuming Zhang, Zhoutong Zhang, Richard Zhang, Jun-Yan Zhu, and Bryan Russell.
\newblock Editing conditional radiance fields.
\newblock In \emph{Proceedings of the IEEE/CVF international conference on computer vision}, pages 5773--5783, 2021.

\bibitem[Liu et~al.(2022)Liu, Peng, Liu, Wang, Wang, Theobalt, Zhou, and Wang]{liu2022neural}
Yuan Liu, Sida Peng, Lingjie Liu, Qianqian Wang, Peng Wang, Christian Theobalt, Xiaowei Zhou, and Wenping Wang.
\newblock Neural rays for occlusion-aware image-based rendering.
\newblock In \emph{Proceedings of the IEEE/CVF Conference on Computer Vision and Pattern Recognition}, pages 7824--7833, 2022.

\bibitem[Luo et~al.(2019{\natexlab{a}})Luo, Guan, Ju, Huang, and Luo]{luo2019pmvsnet}
Keyang Luo, Tao Guan, Lili Ju, Haipeng Huang, and Yawei Luo.
\newblock P-mvsnet: Learning patch-wise matching confidence aggregation for multi-view stereo.
\newblock In \emph{ICCV}, pages 10452--10461, 2019{\natexlab{a}}.

\bibitem[Luo et~al.(2020{\natexlab{a}})Luo, Guan, Ju, Wang, Chen, and Luo]{luo2020attention}
Keyang Luo, Tao Guan, Lili Ju, Yuesong Wang, Zhuo Chen, and Yawei Luo.
\newblock Attention-aware multi-view stereo.
\newblock In \emph{CVPR}, pages 1590--1599, 2020{\natexlab{a}}.

\bibitem[Luo and Yang(2024)]{luo2024LDMC}
Yawei Luo and Yi Yang.
\newblock Large language model and domain-specific model collaboration for smart education.
\newblock \emph{FITEE}, 2024.

\bibitem[Luo et~al.(2019{\natexlab{b}})Luo, Liu, Guan, Yu, and Yang]{luo2019significance}
Yawei Luo, Ping Liu, Tao Guan, Junqing Yu, and Yi Yang.
\newblock Significance-aware information bottleneck for domain adaptive semantic segmentation.
\newblock In \emph{ICCV}, pages 6778--6787, 2019{\natexlab{b}}.

\bibitem[Luo et~al.(2019{\natexlab{c}})Luo, Zheng, Guan, Yu, and Yang]{luo2019taking}
Yawei Luo, Liang Zheng, Tao Guan, Junqing Yu, and Yi Yang.
\newblock Taking a closer look at domain shift: Category-level adversaries for semantics consistent domain adaptation.
\newblock In \emph{CVPR}, pages 2507--2516, 2019{\natexlab{c}}.

\bibitem[Luo et~al.(2020{\natexlab{b}})Luo, Liu, Guan, Yu, and Yang]{luo2020adversarial_NIPS2020}
Yawei Luo, Ping Liu, Tao Guan, Junqing Yu, and Yi Yang.
\newblock Adversarial style mining for one-shot unsupervised domain adaptation.
\newblock In \emph{NeurIPS}, pages 20612--20623, 2020{\natexlab{b}}.

\bibitem[Luo et~al.(2021)Luo, Liu, Zheng, Guan, Yu, and Yang]{luo2021category}
Yawei Luo, Ping Liu, Liang Zheng, Tao Guan, Junqing Yu, and Yi Yang.
\newblock Category-level adversarial adaptation for semantic segmentation using purified features.
\newblock \emph{T-PAMI}, 2021.

\bibitem[Ma et~al.(2023)Ma, Luo, and Yang]{ma2023personas}
Shaojie Ma, Yawei Luo, and Yi Yang.
\newblock Personas-based student grouping using reinforcement learning and linear programming.
\newblock \emph{Knowledge-Based Systems}, page 111071, 2023.

\bibitem[Mildenhall et~al.(2019)Mildenhall, Srinivasan, Ortiz-Cayon, Kalantari, Ramamoorthi, Ng, and Kar]{mildenhall2019local}
Ben Mildenhall, Pratul~P Srinivasan, Rodrigo Ortiz-Cayon, Nima~Khademi Kalantari, Ravi Ramamoorthi, Ren Ng, and Abhishek Kar.
\newblock Local light field fusion: Practical view synthesis with prescriptive sampling guidelines.
\newblock \emph{ACM Transactions on Graphics (TOG)}, 38\penalty0 (4):\penalty0 1--14, 2019.

\bibitem[Mildenhall et~al.(2021)Mildenhall, Srinivasan, Tancik, Barron, Ramamoorthi, and Ng]{mildenhall2021nerf}
Ben Mildenhall, Pratul~P Srinivasan, Matthew Tancik, Jonathan~T Barron, Ravi Ramamoorthi, and Ren Ng.
\newblock Nerf: Representing scenes as neural radiance fields for view synthesis.
\newblock \emph{Communications of the ACM}, 65\penalty0 (1):\penalty0 99--106, 2021.

\bibitem[Molaei et~al.(2023)Molaei, Aminimehr, Tavakoli, Kazerouni, Azad, Azad, and Merhof]{molaei2023implicit}
Amirali Molaei, Amirhossein Aminimehr, Armin Tavakoli, Amirhossein Kazerouni, Bobby Azad, Reza Azad, and Dorit Merhof.
\newblock Implicit neural representation in medical imaging: A comparative survey.
\newblock In \emph{ICCV}, pages 2381--2391, 2023.

\bibitem[Park et~al.(2021)Park, Sinha, Barron, Bouaziz, Goldman, Seitz, and Martin-Brualla]{park2021nerfies}
Keunhong Park, Utkarsh Sinha, Jonathan~T Barron, Sofien Bouaziz, Dan~B Goldman, Steven~M Seitz, and Ricardo Martin-Brualla.
\newblock Nerfies: Deformable neural radiance fields.
\newblock In \emph{Proceedings of the IEEE/CVF International Conference on Computer Vision}, pages 5865--5874, 2021.

\bibitem[Pumarola et~al.(2021)Pumarola, Corona, Pons-Moll, and Moreno-Noguer]{pumarola2021d}
Albert Pumarola, Enric Corona, Gerard Pons-Moll, and Francesc Moreno-Noguer.
\newblock D-nerf: Neural radiance fields for dynamic scenes.
\newblock In \emph{Proceedings of the IEEE/CVF Conference on Computer Vision and Pattern Recognition}, pages 10318--10327, 2021.

\bibitem[Reiser et~al.(2023)Reiser, Szeliski, Verbin, Srinivasan, Mildenhall, Geiger, Barron, and Hedman]{reiser2023merf}
Christian Reiser, Rick Szeliski, Dor Verbin, Pratul Srinivasan, Ben Mildenhall, Andreas Geiger, Jon Barron, and Peter Hedman.
\newblock Merf: Memory-efficient radiance fields for real-time view synthesis in unbounded scenes.
\newblock \emph{ACM Transactions on Graphics (TOG)}, 42\penalty0 (4):\penalty0 1--12, 2023.

\bibitem[Reizenstein et~al.(2021)Reizenstein, Shapovalov, Henzler, Sbordone, Labatut, and Novotny]{reizenstein2021common}
Jeremy Reizenstein, Roman Shapovalov, Philipp Henzler, Luca Sbordone, Patrick Labatut, and David Novotny.
\newblock Common objects in 3d: Large-scale learning and evaluation of real-life 3d category reconstruction.
\newblock In \emph{Proceedings of the IEEE/CVF International Conference on Computer Vision}, pages 10901--10911, 2021.

\bibitem[Ronneberger et~al.(2015)Ronneberger, Fischer, and Brox]{ronneberger2015u}
Olaf Ronneberger, Philipp Fischer, and Thomas Brox.
\newblock U-net: Convolutional networks for biomedical image segmentation.
\newblock In \emph{Medical Image Computing and Computer-Assisted Intervention--MICCAI 2015: 18th International Conference, Munich, Germany, October 5-9, 2015, Proceedings, Part III 18}, pages 234--241. Springer, 2015.

\bibitem[Simonyan and Zisserman(2014)]{simonyan2014two}
Karen Simonyan and Andrew Zisserman.
\newblock Two-stream convolutional networks for action recognition in videos.
\newblock \emph{Advances in neural information processing systems}, 27, 2014.

\bibitem[Suhail et~al.(2022{\natexlab{a}})Suhail, Esteves, Sigal, and Makadia]{suhail2022generalizable}
Mohammed Suhail, Carlos Esteves, Leonid Sigal, and Ameesh Makadia.
\newblock Generalizable patch-based neural rendering.
\newblock In \emph{European Conference on Computer Vision}, pages 156--174. Springer, 2022{\natexlab{a}}.

\bibitem[Suhail et~al.(2022{\natexlab{b}})Suhail, Esteves, Sigal, and Makadia]{suhail2022light}
Mohammed Suhail, Carlos Esteves, Leonid Sigal, and Ameesh Makadia.
\newblock Light field neural rendering.
\newblock In \emph{Proceedings of the IEEE/CVF Conference on Computer Vision and Pattern Recognition}, pages 8269--8279, 2022{\natexlab{b}}.

\bibitem[Sun et~al.(2022{\natexlab{a}})Sun, Sun, and Chen]{sun2022direct}
Cheng Sun, Min Sun, and Hwann-Tzong Chen.
\newblock Direct voxel grid optimization: Super-fast convergence for radiance fields reconstruction.
\newblock In \emph{Proceedings of the IEEE/CVF Conference on Computer Vision and Pattern Recognition}, pages 5459--5469, 2022{\natexlab{a}}.

\bibitem[Sun et~al.(2022{\natexlab{b}})Sun, Wang, Zhang, Li, Zhang, Liu, and Wang]{sun2022fenerf}
Jingxiang Sun, Xuan Wang, Yong Zhang, Xiaoyu Li, Qi Zhang, Yebin Liu, and Jue Wang.
\newblock Fenerf: Face editing in neural radiance fields.
\newblock In \emph{Proceedings of the IEEE/CVF Conference on Computer Vision and Pattern Recognition}, pages 7672--7682, 2022{\natexlab{b}}.

\bibitem[Tian et~al.(2023)Tian, Du, and Duan]{tian2023mononerf}
Fengrui Tian, Shaoyi Du, and Yueqi Duan.
\newblock Mononerf: Learning a generalizable dynamic radiance field from monocular videos.
\newblock In \emph{Proceedings of the IEEE/CVF International Conference on Computer Vision}, pages 17903--17913, 2023.

\bibitem[Tretschk et~al.(2021)Tretschk, Tewari, Golyanik, Zollh{\"o}fer, Lassner, and Theobalt]{tretschk2021non}
Edgar Tretschk, Ayush Tewari, Vladislav Golyanik, Michael Zollh{\"o}fer, Christoph Lassner, and Christian Theobalt.
\newblock Non-rigid neural radiance fields: Reconstruction and novel view synthesis of a dynamic scene from monocular video.
\newblock In \emph{Proceedings of the IEEE/CVF International Conference on Computer Vision}, pages 12959--12970, 2021.

\bibitem[Varma et~al.(2022)Varma, Wang, Chen, Chen, Venugopalan, and Wang]{varma2022attention}
Mukund Varma, Peihao Wang, Xuxi Chen, Tianlong Chen, Subhashini Venugopalan, and Zhangyang Wang.
\newblock Is attention all that nerf needs?
\newblock In \emph{The Eleventh International Conference on Learning Representations}, 2022.

\bibitem[Wang et~al.(2022{\natexlab{a}})Wang, Chai, He, Chen, and Liao]{wang2022clip}
Can Wang, Menglei Chai, Mingming He, Dongdong Chen, and Jing Liao.
\newblock Clip-nerf: Text-and-image driven manipulation of neural radiance fields.
\newblock In \emph{Proceedings of the IEEE/CVF Conference on Computer Vision and Pattern Recognition}, pages 3835--3844, 2022{\natexlab{a}}.

\bibitem[Wang et~al.(2022{\natexlab{b}})Wang, Cui, Salcudean, and Wang]{wang2022generalizable}
Dan Wang, Xinrui Cui, Septimiu Salcudean, and Z~Jane Wang.
\newblock Generalizable neural radiance fields for novel view synthesis with transformer.
\newblock \emph{arXiv preprint arXiv:2206.05375}, 2022{\natexlab{b}}.

\bibitem[Wang et~al.(2021)Wang, Wang, Genova, Srinivasan, Zhou, Barron, Martin-Brualla, Snavely, and Funkhouser]{wang2021ibrnet}
Qianqian Wang, Zhicheng Wang, Kyle Genova, Pratul~P Srinivasan, Howard Zhou, Jonathan~T Barron, Ricardo Martin-Brualla, Noah Snavely, and Thomas Funkhouser.
\newblock Ibrnet: Learning multi-view image-based rendering.
\newblock In \emph{Proceedings of the IEEE/CVF Conference on Computer Vision and Pattern Recognition}, pages 4690--4699, 2021.

\bibitem[Wang et~al.(2022{\natexlab{c}})Wang, Chen, Wu, Chen, Dai, Liu, Jiang, Zhou, and Yuan]{wang2022bevt}
Rui Wang, Dongdong Chen, Zuxuan Wu, Yinpeng Chen, Xiyang Dai, Mengchen Liu, Yu-Gang Jiang, Luowei Zhou, and Lu Yuan.
\newblock Bevt: Bert pretraining of video transformers.
\newblock In \emph{Proceedings of the IEEE/CVF Conference on Computer Vision and Pattern Recognition}, pages 14733--14743, 2022{\natexlab{c}}.

\bibitem[Wang et~al.(2020)Wang, Guan, Chen, Luo, Luo, and Ju]{wang2020mesh}
Yuesong Wang, Tao Guan, Zhuo Chen, Yawei Luo, Keyang Luo, and Lili Ju.
\newblock Mesh-guided multi-view stereo with pyramid architecture.
\newblock In \emph{CVPR}, pages 2039--2048, 2020.

\bibitem[Wang et~al.(2023)Wang, Zeng, Guan, Yang, Chen, Liu, Xu, and Luo]{wang2023adaptive}
Yuesong Wang, Zhaojie Zeng, Tao Guan, Wei Yang, Zhuo Chen, Wenkai Liu, Luoyuan Xu, and Yawei Luo.
\newblock Adaptive patch deformation for textureless-resilient multi-view stereo.
\newblock In \emph{CVPR}, pages 1621--1630, 2023.

\bibitem[Wizadwongsa et~al.(2021)Wizadwongsa, Phongthawee, Yenphraphai, and Suwajanakorn]{wizadwongsa2021nex}
Suttisak Wizadwongsa, Pakkapon Phongthawee, Jiraphon Yenphraphai, and Supasorn Suwajanakorn.
\newblock Nex: Real-time view synthesis with neural basis expansion.
\newblock In \emph{Proceedings of the IEEE/CVF Conference on Computer Vision and Pattern Recognition}, pages 8534--8543, 2021.

\bibitem[Xiang et~al.(2021)Xiang, Xu, Hasan, Hold-Geoffroy, Sunkavalli, and Su]{xiang2021neutex}
Fanbo Xiang, Zexiang Xu, Milos Hasan, Yannick Hold-Geoffroy, Kalyan Sunkavalli, and Hao Su.
\newblock Neutex: Neural texture mapping for volumetric neural rendering.
\newblock In \emph{Proceedings of the IEEE/CVF Conference on Computer Vision and Pattern Recognition}, pages 7119--7128, 2021.

\bibitem[Xu and Harada(2022)]{xu2022deforming}
Tianhan Xu and Tatsuya Harada.
\newblock Deforming radiance fields with cages.
\newblock In \emph{European Conference on Computer Vision}, pages 159--175. Springer, 2022.

\bibitem[Yang et~al.(2023)Yang, Hong, Li, Hu, Li, Lee, and Wang]{yang2023contranerf}
Hao Yang, Lanqing Hong, Aoxue Li, Tianyang Hu, Zhenguo Li, Gim~Hee Lee, and Liwei Wang.
\newblock Contranerf: Generalizable neural radiance fields for synthetic-to-real novel view synthesis via contrastive learning.
\newblock In \emph{Proceedings of the IEEE/CVF Conference on Computer Vision and Pattern Recognition}, pages 16508--16517, 2023.

\bibitem[Yin et~al.(2022{\natexlab{a}})Yin, Fang, Zhou, Zhang, Xu, Shen, and Wang]{yin2022semi}
Junbo Yin, Jin Fang, Dingfu Zhou, Liangjun Zhang, Cheng-Zhong Xu, Jianbing Shen, and Wenguan Wang.
\newblock Semi-supervised 3d object detection with proficient teachers.
\newblock In \emph{ECCV}, pages 727--743, 2022{\natexlab{a}}.

\bibitem[Yin et~al.(2022{\natexlab{b}})Yin, Zhou, Zhang, Fang, Xu, Shen, and Wang]{yin2022proposalcontrast}
Junbo Yin, Dingfu Zhou, Liangjun Zhang, Jin Fang, Cheng-Zhong Xu, Jianbing Shen, and Wenguan Wang.
\newblock Proposalcontrast: Unsupervised pre-training for lidar-based 3d object detection.
\newblock In \emph{ECCV}, pages 17--33, 2022{\natexlab{b}}.

\bibitem[Yu et~al.(2021{\natexlab{a}})Yu, Li, Tancik, Li, Ng, and Kanazawa]{yu2021plenoctrees}
Alex Yu, Ruilong Li, Matthew Tancik, Hao Li, Ren Ng, and Angjoo Kanazawa.
\newblock Plenoctrees for real-time rendering of neural radiance fields.
\newblock In \emph{Proceedings of the IEEE/CVF International Conference on Computer Vision}, pages 5752--5761, 2021{\natexlab{a}}.

\bibitem[Yu et~al.(2021{\natexlab{b}})Yu, Ye, Tancik, and Kanazawa]{yu2021pixelnerf}
Alex Yu, Vickie Ye, Matthew Tancik, and Angjoo Kanazawa.
\newblock pixelnerf: Neural radiance fields from one or few images.
\newblock In \emph{Proceedings of the IEEE/CVF Conference on Computer Vision and Pattern Recognition}, pages 4578--4587, 2021{\natexlab{b}}.

\bibitem[Zhang et~al.(2020)Zhang, Riegler, Snavely, and Koltun]{zhang2020nerf++}
Kai Zhang, Gernot Riegler, Noah Snavely, and Vladlen Koltun.
\newblock Nerf++: Analyzing and improving neural radiance fields.
\newblock \emph{arXiv preprint arXiv:2010.07492}, 2020.

\bibitem[Zheng et~al.(2022)Zheng, Huang, Yu, Zhang, Guo, and Liu]{zheng2022structured}
Zerong Zheng, Han Huang, Tao Yu, Hongwen Zhang, Yandong Guo, and Yebin Liu.
\newblock Structured local radiance fields for human avatar modeling.
\newblock In \emph{Proceedings of the IEEE/CVF Conference on Computer Vision and Pattern Recognition}, pages 15893--15903, 2022.

\bibitem[Zhou et~al.(2018)Zhou, Tucker, Flynn, Fyffe, and Snavely]{zhou2018stereo}
Tinghui Zhou, Richard Tucker, John Flynn, Graham Fyffe, and Noah Snavely.
\newblock Stereo magnification: Learning view synthesis using multiplane images.
\newblock \emph{arXiv preprint arXiv:1805.09817}, 2018.

\end{thebibliography}
}

\clearpage \appendix 

\section{Implementation Details}
\subsection{Generalizable 3D Representation $\boldsymbol{z}$}
EVE-NeRF alternates the aggregation of epipolar and view dimensions' features through the VEI and EVI modules, resulting in the generation of generalizable 3D representation $\boldsymbol{z}$ that aligns with NeRF's coordinates.
The pseudocode for the computation of $\boldsymbol{z}$ is as follows:
\begin{algorithm}
  \caption{EVE-NeRF:PyTorch-like Pseudocode}\label{algorithm1}
  \KwIn{viewpoints difference $\Delta \boldsymbol{d}^{s}$, 
  extracted convolution features $\boldsymbol{f}^{c}\in \mathbb{R}^{N\times M\times C}$, numbers of aggregation module $N_{layer}$}
  \KwOut{generalizable 3D representation $\boldsymbol{z}$ }

  $\boldsymbol{X}= \boldsymbol{f}^{c}$\;
  $i=1$\;
\While{$i\leq N_{layer}$}{
$\boldsymbol{h} = \boldsymbol{X}$\;
$\boldsymbol{Q}=\boldsymbol{X}\boldsymbol{W_{Q}},\boldsymbol{K}=\boldsymbol{X}\boldsymbol{W_{K}},\boldsymbol{V}=\boldsymbol{X}\boldsymbol{W_{V}}$\;
$\boldsymbol{X}=\text{VEI}\left(\boldsymbol{Q}, \boldsymbol{K},\boldsymbol{V},\Delta \boldsymbol{d}^{s}\right)$\;
$\boldsymbol{Mean},\boldsymbol{Var} =\text{mean\&var}\left(\boldsymbol{V}, \text{dim}=1\right)$\;
$\boldsymbol{w}^{v} = \text{sigmoid}\left(\text{AE}\left(\boldsymbol{Mean}, \boldsymbol{Var}\right)\right)$\;
$\boldsymbol{X} = \boldsymbol{X}\cdot \boldsymbol{w}^{v}$\;
$\boldsymbol{X}'=\boldsymbol{X.}\text{permute}\left(1, 0, 2\right)$\;
$\boldsymbol{Q}'=\boldsymbol{X}'\boldsymbol{W'_{Q}},\boldsymbol{K}'=\boldsymbol{X}'\boldsymbol{W'_{K}},\boldsymbol{V}'=\boldsymbol{X}'\boldsymbol{W'_{V}}$\;
$\boldsymbol{X}'=\text{EVI}\left(\boldsymbol{Q}', \boldsymbol{K}',\boldsymbol{V}',\Delta \boldsymbol{d}^{s}\right)$\;
$\boldsymbol{Max} =\text{max}\left(\boldsymbol{V}', \text{dim}=1\right)$\;
$\boldsymbol{w}^{e} = \text{sigmoid}\left(\text{self-attn}\left(\boldsymbol{Max}\right)\right)$\;
$\boldsymbol{X}' = \boldsymbol{X}'\cdot \boldsymbol{w}^{e}$\;
$\boldsymbol{X}=\boldsymbol{X}'.\text{permute}\left(1, 0, 2\right)+\boldsymbol{h}$\;
$ i=i+1$\;
}
$\boldsymbol{z}= \text{mean}(\boldsymbol{X}, \text{dim}=1) \in \mathbb{R}^{N\times C}$\;

\end{algorithm}

\subsection{Difference of Views $\Delta\boldsymbol{d}^{s}$}
\label{Difference of Views}
$\Delta\boldsymbol{d}^{s}$  serves as an additional input to attention computation in the view transformer and the epipolar transformer, allowing the model to learn more information about the differences in views. The pseudo-code for computing $\Delta\boldsymbol{d}^{s}$ is shown in Algorithm \ref{algorithm2}.

\begin{algorithm}
  \caption{$\Delta\boldsymbol{d}^{s}$:PyTorch-like Pseudocode}\label{algorithm2}
  \KwIn{the target ray direction $\boldsymbol{d}_{t}\in \mathbb{R}^{3}$, the source ray direction $\boldsymbol{d}_{s}\in \mathbb{R}^{M\times 3}$ , the number of sampling points along the target ray $N$}
  \KwOut{$\Delta\boldsymbol{d}^{s}$}

$\boldsymbol{d}_{t} = \boldsymbol{d}_{t}.\text{unsqueeze}(0).\text{repeat}(M, 1)$\;
$\boldsymbol{d}_{diff} = \boldsymbol{d}_{t} - \boldsymbol{d}_{s}$\;
$\boldsymbol{d}_{diff} = \boldsymbol{d}_{diff}/\text{torch.norm}(\boldsymbol{d}_{diff}, \text{dim}=-1, \text{keepdim=True})$\;
$\boldsymbol{d}_{dot} = \text{torch.sum}(\boldsymbol{d}_{t}*\boldsymbol{d}_{s}$)\;
$\boldsymbol{\Delta\boldsymbol{d}^{s}} = \text{torch.cat}([\boldsymbol{d}_{diff}, \boldsymbol{d}_{dot}], \text{dim}=-1)$\;
$\boldsymbol{\Delta\boldsymbol{d}^{s}} = \boldsymbol{\Delta\boldsymbol{d}^{s}}.\text{unsqueeze}(0).\text{repeat}(N, 1, 1) \in \mathbb{R}^{N\times M \times 4}$\;

\end{algorithm}

\subsection{Difference of Camera Poses $\Delta\boldsymbol{pose}$}
\label{Difference of Camera Poses}
$\Delta\boldsymbol{pose}$   provides camera disparity information for multi-view calibration, which is merged with epipolar aggregation features to obtain geometry consistency prior. The pseudo-code to compute $\Delta\boldsymbol{pose}$ is shown in Algorithm \ref{algorithm3}.

\begin{algorithm}
  \caption{$\Delta\boldsymbol{pose}$:PyTorch-like Pseudocode}\label{algorithm3}
  \KwIn{the target pose matrix $\boldsymbol{P}_{t}\in \mathbb{R}^{3\times 4}$, the source pose matrix $\boldsymbol{P}_{s}\in \mathbb{R}^{M\times 3\times 4}$}
  \KwOut{$\Delta\boldsymbol{pose}$}

$M=\boldsymbol{P}_{s}.shape[0]$\;
$\boldsymbol{P}_{t} = \boldsymbol{P}_{t}$.unsqueeze(dim=0).repeat($M,1,1$)\;
$\boldsymbol{R}_{t}=\boldsymbol{P}_{t}[:,:3,:3]$\;
$\boldsymbol{R}_{s}=\boldsymbol{P}_{s}[:,:3,:3]$\;
$\boldsymbol{T}_{t}=\boldsymbol{P}_{t}[:,:3,-1]$\;
$\boldsymbol{T}_{s}=\boldsymbol{P}_{s}[:,:3,-1]$\;
$\Delta\boldsymbol{R}=\boldsymbol{R}_{t}\text{@}\boldsymbol{R}_{s}^{T}.\text{view}(M, 9)$\;
$\Delta\boldsymbol{T}=\boldsymbol{T}_{t}-\boldsymbol{T}_{s}^{T}$\;
$\Delta\boldsymbol{pose} = \text{torch.cat}([\Delta\boldsymbol{R}, \Delta\boldsymbol{T}], \text{dim=-1}) \in \mathbb{R}^{M\times 12}$\;

\end{algorithm}

\subsection{Additional Technical Details}
\label{Additional Technical Details}
\textbf{EVE-NeRF network details.}\; Our lightweight CNN consists of 4 convolutional layers with a kernel size of $3\times 3$ and a stride of 1. BatchNorm layers and ReLU activation functions are applied between layers. The final output feature map has a dimension of 32.
The VEI and EVI modules have 4 layers, which are connected alternately. Both the View Transformer and Epipolar Transformer have the same network structure, in which the dimension of hidden features is 64 and we use 4 heads for the self-attention module in transformer layers. For the transformer in Multi-View Calibration, the features dimension is 64 and head is 4, consisting of 1 blocks. For the AE  network in Along-Epipolar Perception and the conditioned NeRF decoder are set the same as the experimental setups of GeoNeRF~\cite{johari2022geonerf} and IBRNet~\cite{wang2021ibrnet}, respectively.
The network architectures of the lightweight CNN, the AE network, and the conditioned NeRF decoder are provided in Table \ref{tab:light-weight-cnn}, \ref{tab:conv1d-AE}, and \ref{tab:nerf-decoder} respectively.\vspace{8pt}
\begin{table}[t]
\centering
\resizebox{\columnwidth}{!}{

\begin{tabular}{ccc}
\toprule 
Input & Layer  & Output  \\           
\midrule
input & Conv2d(3, 32, 3, 1)+BN+ReLU & conv0  \\
conv0 & Conv2d(32, 32, 3, 1)+BN+ReLU & conv1  \\
conv1 & Conv2d(32, 32, 3, 1)+BN & conv$2\_0$  \\
(conv0, conv$2\_0$) & Add(conv0, conv$2\_0$) + ReLU & conv$2\_1$  \\
conv$2\_1$ & Conv2d(32, 32, 3, 1)+BN+ReLU & \textbf{conv3} \\
\bottomrule
\end{tabular}

}
\caption{
Network architecture of the lightweight CNN, where $\textbf{conv3}$ is the output features. Conv2d($c_{in}$, $c_{out}$, $k$, $s$) stands for a 2D convolution with input channels $c_{in}$, output channels $c_{out}$, kernel size of $k$, and stride of $s$. BN stands for Batch Normalization Layer. ReLU stands for ReLU nonlinearity activation function. Add(x, y) means add x and y.
}
\vspace{-0.05in}
\label{tab:light-weight-cnn}
\end{table}
\begin{table}[t]
\centering
\resizebox{\columnwidth}{!}{

\begin{tabular}{ccc}
\toprule 
Input & Layer  & Output  \\           
\midrule
input & Conv1d(128, 64, 3, 1)+LN+ELU & conv$1\_0$  \\
conv$1\_0$ & MaxPool1d & conv1  \\
conv1 & Conv1d(64, 128, 3, 1)+LN+ELU & conv$2\_0$  \\
conv$2\_0$ & MaxPool1d & conv2  \\
conv2 & Conv1d(128, 128, 3, 1)+LN+ELU & conv$3\_0$  \\
conv$3\_0$ & MaxPool1d & conv3  \\
conv3 & TrpsConv1d(128, 128, 4, 2)+LN+ELU & x$\_$0  \\
$[$conv2;x$\_$0$]$ & TrpsConv1d(256, 64, 4, 2)+LN+ELU & x$\_$1  \\
$[$conv1;x$\_$1$]$ & TrpsConv1d(128, 32, 4, 2)+LN+ELU & x$\_$2  \\
$[$Input;x$\_$2$]$ & Conv1d(64, 64, 3, 1)+Sigmoid & \textbf{output}  \\
\bottomrule
\end{tabular}

}
\caption{
Network architecture of the 1D convolution AE. Conv2d($c_{in}$, $c_{out}$, $k$, $s$) stands for a 1D convolution with input channels $c_{in}$, output channels $c_{out}$, kernel size of $k$, and stride of $s$. LN stands for Layer Normalization Layer. ELU and Sigmoid stand for ELU and Sigmoid nonlinearity activation function separately. MaxPool1d is a 1D max pooling layer with a stride of 2. TrpsConv1d stands for transposed 1D convolution. $\left[\cdot;\cdot \right]$ means concatenation.
}
\vspace{-0.05in}
\label{tab:conv1d-AE}
\end{table}
\begin{table}[t]
\centering

\begin{tabular}{ccc}
\toprule 
Input & Layer  & Output  \\           
\midrule
$\boldsymbol{z}$ & Linear(64, 128) & bias  \\
$\gamma(\boldsymbol{p})$ & Linear(63, 128) & x0$\_$0  \\
x0$\_$0,bias & Mul(x0$\_$0,bias)+ReLU & x0  \\
x0 & Linear(128, 128) & x1$\_$0  \\
x1$\_$0,bias & Mul(x1$\_$0,bias)+ReLU & x1  \\
x1 & Linear(128, 128) & x2$\_$0  \\
x2$\_$0,bias & Mul(x2$\_$0,bias)+ReLU & x2  \\
x2 & Linear(128, 128) & x3$\_$0  \\
x3$\_$0,bias & Mul(x3$\_$0,bias)+ReLU & x3  \\
x3 & Linear(128, 128) & x4$\_$0  \\
x4$\_$0,bias & Mul(x4$\_$0,bias)+ReLU & x4  \\
$[$x4;$\gamma(\boldsymbol{p})]$ & Linear(191, 128) & x5$\_$0  \\
x5$\_$0,bias & Mul(x5$\_$0,bias)+ReLU & x5  \\
x5 & Linear(128, 16)+ReLU & alpha$\_$raw  \\
alpha$\_$raw & Mul(4, 16) & alpha0  \\
alpha0 & Linear(16,16)+ReLU & alpha1  \\
alpha1 & Linear(16,1)+ReLU & \textbf{alpha}  \\
$[$x5;$\gamma(\boldsymbol{d})]$ & Linear(191,64)+ReLU & x6  \\
x6 &  Linear(64, 3)+Sigmoid  & \textbf{rgb}  \\

\bottomrule
\end{tabular}

\caption{
Network architecture of the conditioned NeRF decoder. $\boldsymbol{z}$, $\boldsymbol{p}$, and $\boldsymbol{d}$ stand for the generalizable features, the coordinates of 3D sampling points, and the directions of rays, individually. $\gamma$ stands for positional encoding in NeRF. Linear($c_{in},c_{out}$) stands for a linear layer with input channels $c_{in}$ and output channels $c_{out}$. Mul stands for element-wise multiplication. MHA($head, dim$) stands for a multi-head-attention  layer with the number of head $head$ and attention dimension $dim$. $\left[\cdot;\cdot \right]$ means concatenation. 
}
\vspace{-0.05in}
\label{tab:nerf-decoder}
\end{table}

\begin{figure*}[htbp]
    \centering
    \begin{subfigure}{0.49\linewidth}
		\centering
		\includegraphics[width=\linewidth]{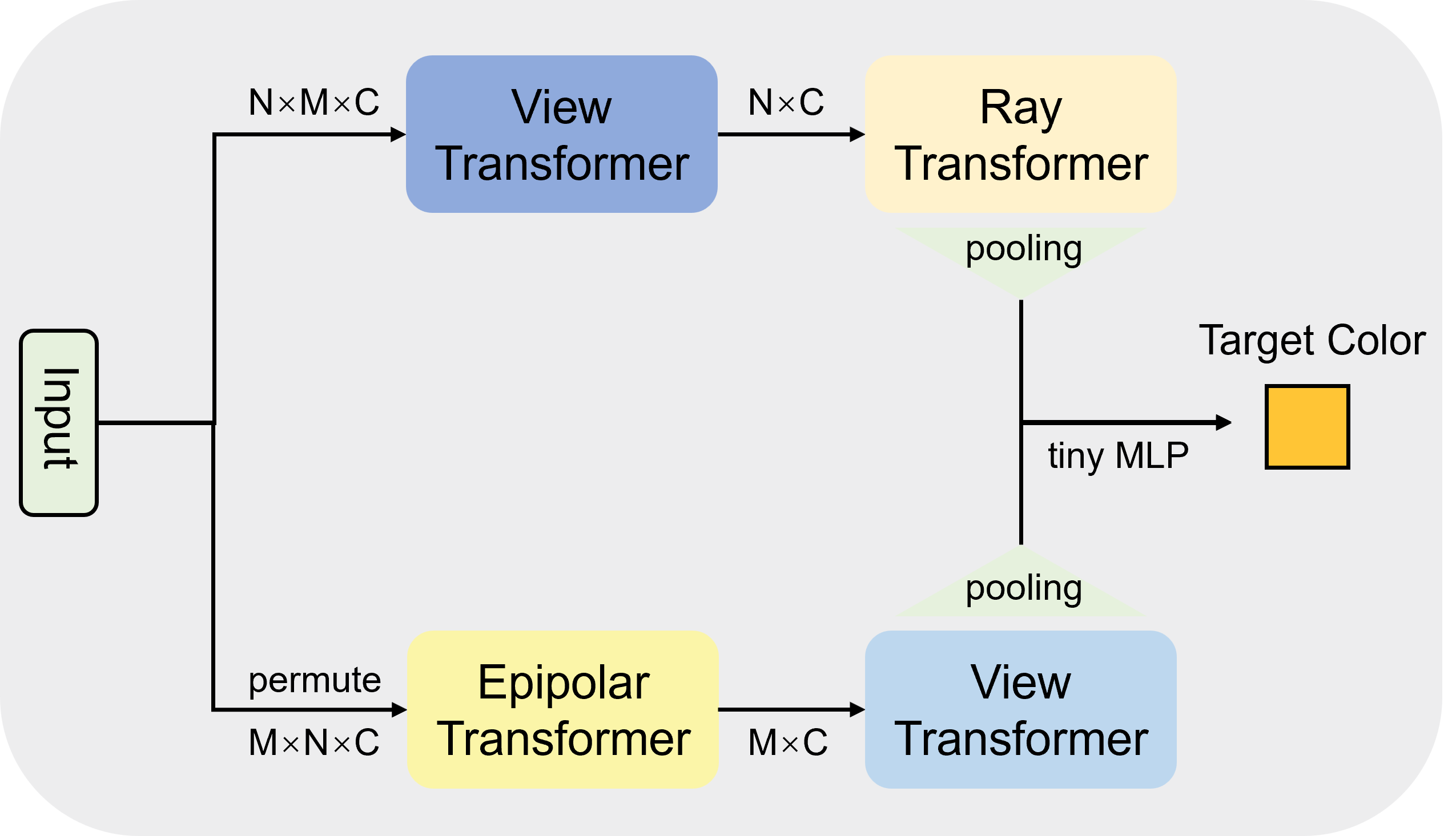}
		\caption{Na\"ive Dual Transformer}
	\end{subfigure}
\centering
 \begin{subfigure}{0.49\linewidth}
		\centering
		\includegraphics[width=\linewidth]{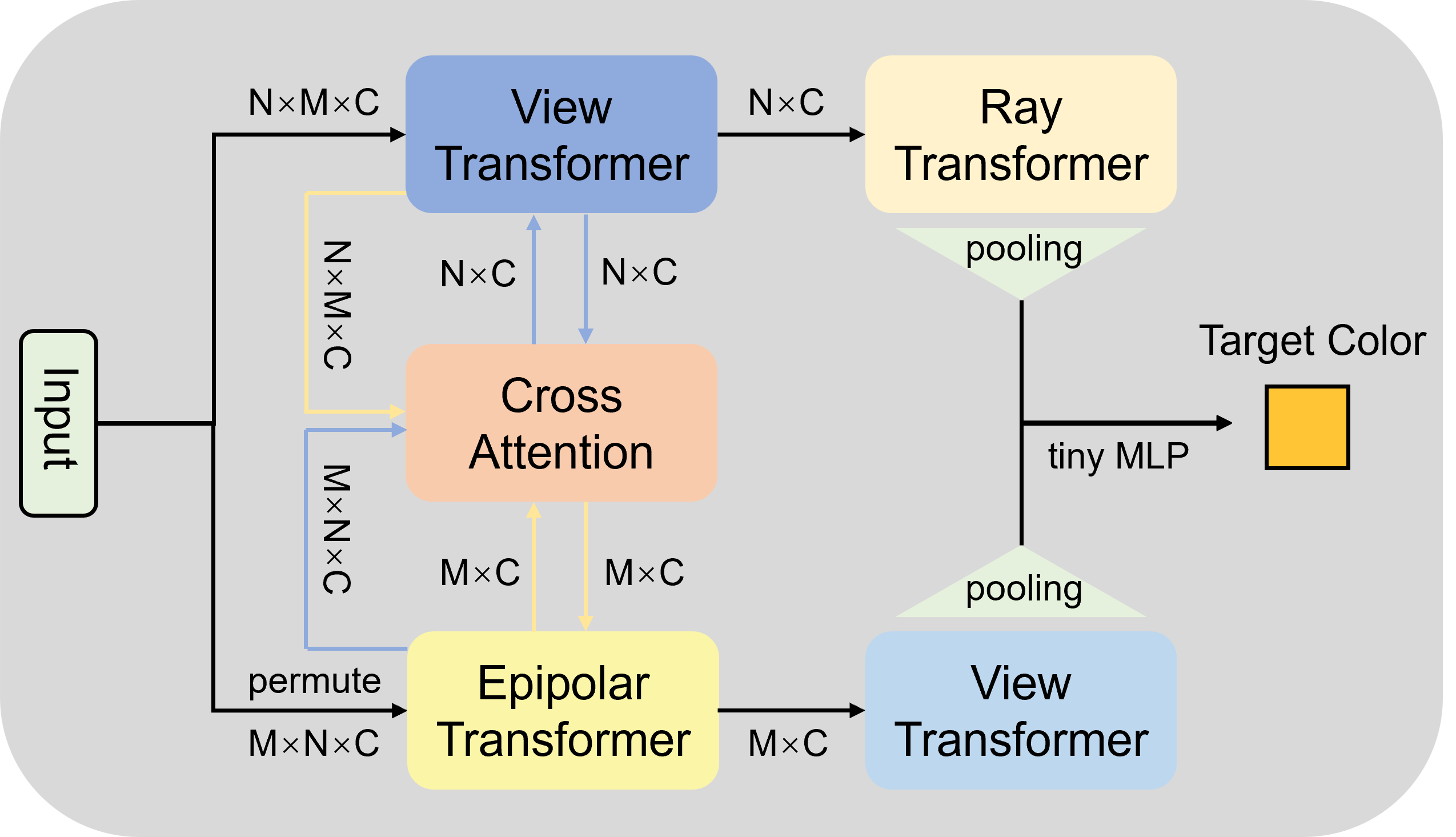}
		\caption{w/ Cross-Attention Interaction}
	\end{subfigure}
 \caption{Na\"ive dual network architecture. We design 2 baselines of dual networks for comparison: a) the Na\"ive Dual Transformer and b) the Dual Transformer with Cross-Attention Interaction. Table \ref{tab:results for ablation} demonstrates that our proposed method, EVE-NeRF, exhibits superior generalization capabilities for novel view synthesis.}
	\label{naive_structure}
\end{figure*}

\noindent\textbf{Na\"ive dual network details}.\; To further validate the rationality of EVE-NeRF's dual-branch structure, in Sec \ref{ablation}, we compared our method with two na\"ive dual network architectures: the Na\"ive Dual Transformer and the Dual Transformer with Cross-Attention Interaction. The Na\"ive Dual Transformer's first branch is GNT~\cite{varma2022attention}, and the second branch is GNT with epipolar aggregation followed by view aggregation. The dual branch predicts colors of each branch via a tiny MLP network directly. And the final color is the average pooling of the two branch colors. GNT demonstrated that using volume rendering to calculate color values does not enhance GNT's performance. Hence, we consider it fair to compare EVE-NeRF with these two dual-branch networks. The Dual Transformer with Cross-Attention Interaction builds upon the Na\"ive Dual Transformer by adding a cross-attention layer for inter-branch interaction. These dual network architectures are illustrated in Figure \ref{naive_structure}.


\section{Multi-View Epipolar-Aligned Feature Extraction}
Let $\boldsymbol{K}_{t}$ and $\boldsymbol{P}_{t}=[\boldsymbol{R}_{t}, \boldsymbol{t}_{t}]$ represent the camera intrinsic and extrinsic parameters for the target view, and let $\boldsymbol{u}_{t}$ be the pixel coordinates corresponding to the target ray $\mathcal{R}$. In this case, $\mathcal{R}$ can be parameterized in the world coordinate system based on the delta parameter as follows:
\begin{gather}
    \mathcal{R}(\delta) = \boldsymbol{t}_{t}+\delta\boldsymbol{R}_{t}\boldsymbol{K}_{t}^{-1}[\boldsymbol{u}_{t}^\top, 1]^\top.
\end{gather}

Next, we sample $N$ points $\{\boldsymbol{p}_{i}\}_{i=1}^{N}=\{\mathcal{R}(\delta_{i})\}_{i=1}^{N}$ along $\mathcal{R}$ and project them onto the $j$-th source view:
\begin{gather}
    d_{j}^{i}[{u_{j}^{i}}^\top, 1]^\top = \boldsymbol{K}_{j}\boldsymbol{R}_{j}^{-1}(\boldsymbol{p}_{i}-\boldsymbol{t}_{j}),
\end{gather}
where $u_{j}^{i}$ is the 2D coordinates of the $i$-th sampled point's projection onto the $j$-th source view, and $d_{j}^{i}$ is the corresponding depth. Clearly, the projection points of these sampled points lie on the corresponding epipolar line in that view. Next, we obtain the convolution features $\boldsymbol{f}^{c}=\{\boldsymbol{f}^{c}_{i,j}\}_{i=1,j=1}^{N,M}$ in $\{\boldsymbol{F}_{i}^{c}\}_{i=1}^{M}$ for these projection points via bilinear interpolation. Therefore, for the target ray $\mathcal{R}$, we now have the multi-view convolution features $\boldsymbol{f}^{c}\in \mathbb{R}^{N\times M\times C}$ for $\mathcal{R}$, where $C$ is the number of channels in the convolution features.


\begin{figure*}[htbp]
\centering
    \begin{subfigure}{0.14\linewidth}
		\centering
		\includegraphics[width=\linewidth]{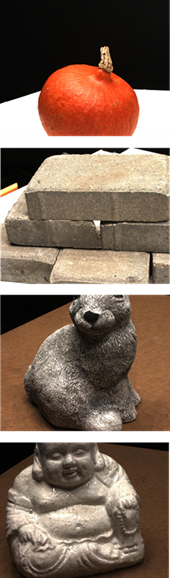}
		\caption{Ground Truth}
	\end{subfigure}
    \centering
    \begin{subfigure}{0.28\linewidth}
		\centering
		\includegraphics[width=\linewidth]{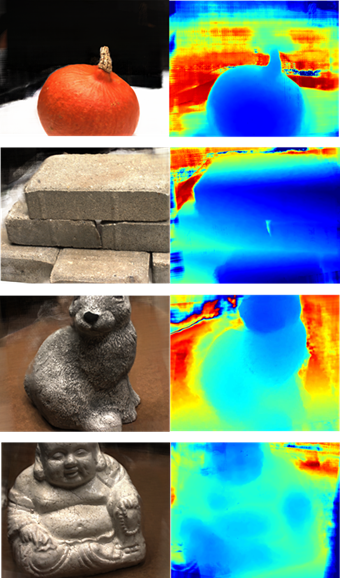}
		\caption{MVSNeRF}
	\end{subfigure}
\centering
 \begin{subfigure}{0.28\linewidth}
		\centering
		\includegraphics[width=\linewidth]{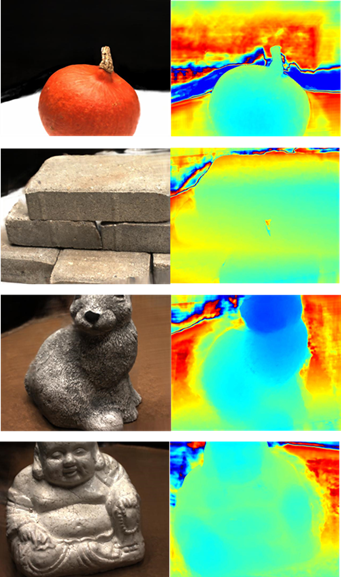}
		\caption{MatchNeRF}
	\end{subfigure}
 \centering
 \begin{subfigure}{0.28\linewidth}
		\centering
		\includegraphics[width=\linewidth]{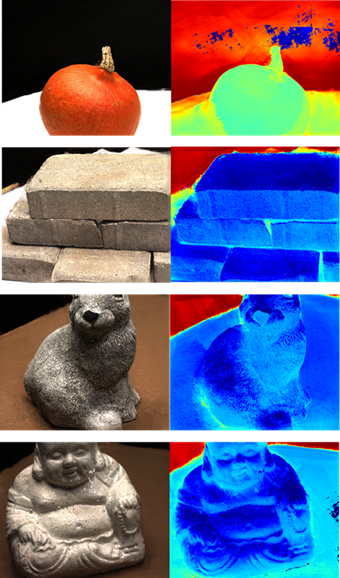}
		\caption{EVE-NeRF}
	\end{subfigure}
 \caption{Qualitative comparison of our generalizable GeoNeRF model with MVSNeRF~\cite{chen2021mvsnerf} and MatchNeRF~\cite{chen2023explicit} in the few-shot setting. Our proposed method, EVE-NeRF, not only has higher rendering of new view pictures but also provides more accurate and detailed depth maps (without ground-truth depth supervision).This is due to the fact that EVE-NeRF provides accurate geometric and appearance a prior of multiple views for the model through the complementary structure of epipolar aggregation and view aggregation.}
	\label{setting2_dtu}
\end{figure*}

 \begin{figure}[t]
    \centering
    \begin{subfigure}{\columnwidth}
		\centering
		\includegraphics[width=\columnwidth]{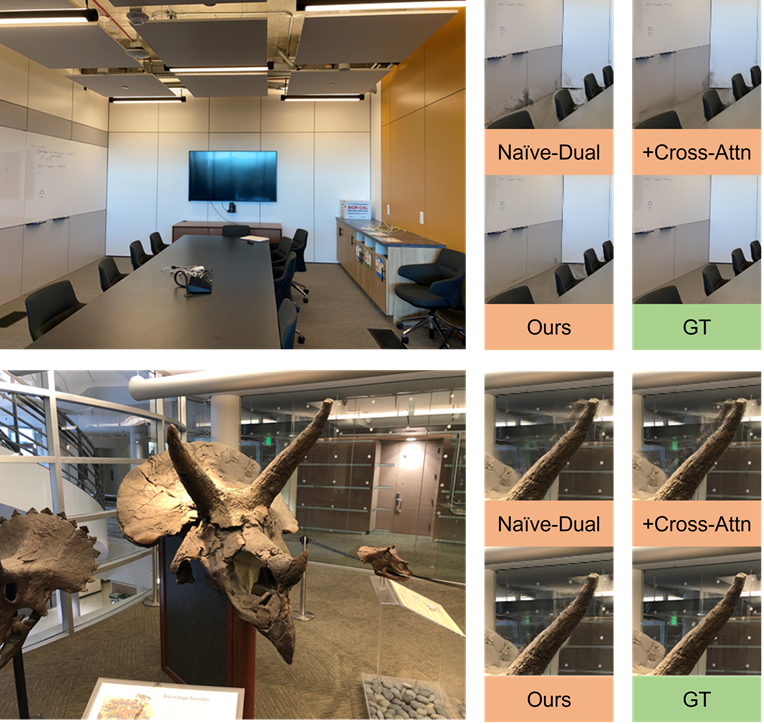}
		\caption{LLFF datasets}
		\label{naive_compare_llff}
	\end{subfigure}

    \begin{subfigure}{\columnwidth}
		\centering
		\includegraphics[width=\columnwidth]{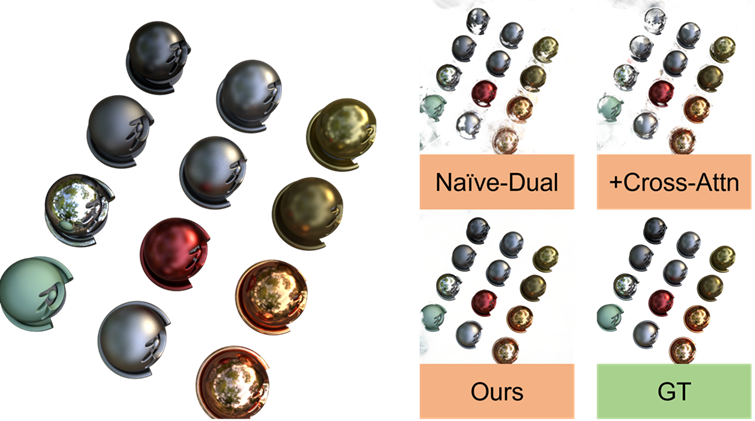}
		\caption{Blender datasets}
		\label{naive_compare_blender}
	\end{subfigure}
 \caption{Qualitative comparison with na\"ive dual network architectures.}
	\label{naive_compare_fig}
\end{figure}

\section{Feature Aggregation Network Proposed in Other Domains}
Dual-branch network structures are commonly used in computer vision tasks~\cite{simonyan2014two,wang2022bevt,chen2021crossvit,chen2023dual}. For instance, Simonyan~\cite{simonyan2014two} introduced a dual-stream network for action recognition in videos, consisting of a temporal stream for optical flow data and a spatial stream for RGB images, with the outputs from both branches being fused in the end. CrossVIT~\cite{chen2021crossvit} is a visual Transformer model based on dual branches, designed to enable the model to learn multi-scale feature information by processing different-sized image patches through the dual-branch network. DAT~\cite{chen2023dual}, on the other hand, is a transformer-based image super-resolution network that aggregates spatial and channel features through alternating spatial window self-attention and channel self-attention, enhancing representation capacity.
Our approach does not follow the naive dual-branch structure. Instead, we introduce the along-epipolar perception and the multi-view calibration to compensate for the shortcomings in information interaction of the other branch. Besides, our dual-branch network demonstrates the efficient interplay between branches. 

\begin{figure}[htbp]
	\centering
	\begin{subfigure}{0.49\linewidth}
		\centering
		\includegraphics[width=\linewidth]{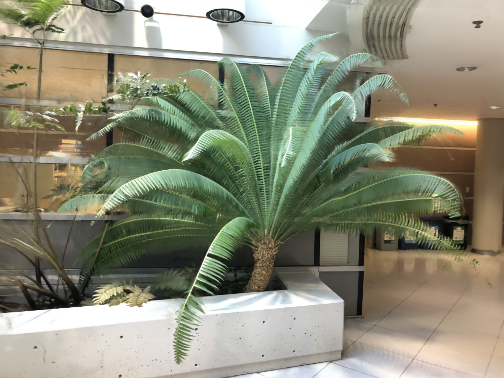}
		\caption{Branch1 Rendering Results}
		\label{Branch1 Rendering Results}
	\end{subfigure}
	\centering
	\begin{subfigure}{0.49\linewidth}
		\centering
		\includegraphics[width=\linewidth]{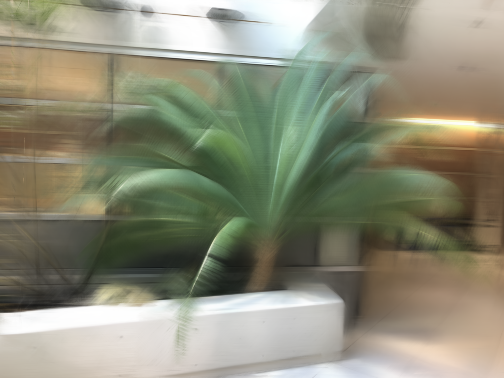}
		\caption{Branch2 Rendering Results}
		\label{Branch2 Rendering Results}
	\end{subfigure}
	
	\caption{Qualitative comparison with dual branches within the Na\"ive Dual Transformer}
	\label{Branch Rendering Results}
\end{figure}

\section{Additional Results}

\subsection{Qualitative Comparison for Setting 2}
\label{Qualitative Comparison for Setting 2}
A qualitative comparison of our method with the few-shot generalizable neural rendering methods~\cite{chen2021mvsnerf,chen2023explicit} is shown in Figure \ref{setting2_dtu}. The novel view images rendered by our method produce minimal artifacts and can render the edge portion of the image and weakly textured regions. 
In addition, we generate a novel view depth map with 3 source views input through the volume rendering~\cite{mildenhall2021nerf}. 
From Figure \ref{setting2_dtu} we can observe that our generated depth map is more accurate and precise in terms of scene geometry prediction. This indicates that our proposed EVE-NeRF can extract high-quality aggregated features that imply the geometry and appearance of the scene, even in a few-shot setting.


\subsection{Per-Scene Fine-Tuning Results}
\begin{table*}[t]
\centering
\begin{subtable}{1.\linewidth}
      \centering
        \begin{tabular}{ccccccccccc}
            \toprule
            Models & Room& Fern&  Leaves & Fortress&  Orchids&  Flower&  T-Rex&  Horns & Avg\\  
            \midrule
            NeRF~\cite{mildenhall2021nerf}  &32.70 &25.17 &20.92 &31.16 &20.36& 27.40 &26.80 &27.45    &  26.50 \\
            NeX~\cite{wizadwongsa2021nex}  &32.32 &\underline{25.63} &21.96 &31.67 &20.42 &28.90 &28.73 &28.46 & 27.26\\
            NLF~\cite{suhail2022light} &\textbf{34.54} &24.86 &\underline{22.47} &\textbf{33.22} &\underline{21.05} &\textbf{29.82} &\textbf{30.34} &\underline{29.78} & \underline{28.26}\\
             EVE-NeRF  &\underline{33.97}    &\textbf{25.73}     &\textbf{23.78}          & \underline{32.97}       &\textbf{21.27}           &\underline{29.06}       &\underline{29.18}     & \textbf{30.53}    & \textbf{28.31 } \\
            \bottomrule
        \end{tabular}
        \caption{PSNR$\uparrow$}
    \end{subtable}%

\begin{subtable}{1.\linewidth}
      \centering
        \begin{tabular}{ccccccccccc}
            \toprule
            Models & Room& Fern&  Leaves & Fortress&  Orchids&  Flower&  T-Rex&  Horns & Avg\\  
            \midrule
            NeRF~\cite{mildenhall2021nerf} &0.948 &0.792 &0.690&0.881 &0.641 &0.827 &0.880 &0.828 &0.811\\
            NeX~\cite{wizadwongsa2021nex} &0.975 &\underline{0.887}&0.832 &0.952& 0.765& 0.933 &0.953 &0.934 &0.904\\
            NLF~\cite{suhail2022light}& \textbf{0.987} &0.886 &\underline{0.856} &\textbf{0.964} &\textbf{0.807}& \textbf{0.939} &\textbf{0.968} &\underline{0.957} &\underline{0.921}\\
             EVE-NeRF  &\underline{0.983}      &\textbf{0.894}     &\textbf{0.891}          & \underline{0.961}       & \underline{0.797}          &\underline{0.935}       &\underline{0.960}     &  \textbf{0.961}     &\textbf{0.923} \\
            \bottomrule
        \end{tabular}
        \caption{SSIM$\uparrow$}
    \end{subtable}%

\begin{subtable}{1.\linewidth}
      \centering
        \begin{tabular}{ccccccccccc}
            \toprule
            Models & Room& Fern&  Leaves & Fortress&  Orchids&  Flower&  T-Rex&  Horns &Avg\\  
            \midrule
            NeRF~\cite{mildenhall2021nerf} &0.178& 0.280& 0.316&0.171 &0.321& 0.219 &0.249&0.268 &0.250\\
            NeX~\cite{wizadwongsa2021nex} &0.161 &0.205 &0.173& 0.131 &0.242 &0.150 &0.192 &0.173 &0.178\\
            NLF~\cite{suhail2022light}& \underline{0.104} &\textbf{0.135}& \textbf{0.110} &\underline{0.119}& \textbf{0.173} &\underline{0.107} &\underline{0.143} &\underline{0.121} &\underline{0.127}\\
             EVE-NeRF  & \textbf{0.060}      &\underline{0.140}     &\underline{0.119}          & \textbf{0.089}       & \underline{0.186}          & \textbf{0.103}      & \textbf{0.095}    & \textbf{0.086}      & \textbf{0.110}\\
            \bottomrule
        \end{tabular}
        \caption{LPIPS$\downarrow$}
    \end{subtable}%

\caption{ Single-scene fine-tuned comparison results for the LLFF dataset
}
\vspace{-0.05in}
\label{tab:finetune_llff}
\end{table*}

We fine-tune for $60,000$ iterations for each scene on the LLFF dataset. The quantitative comparison of our method with single-scene NeRF is demonstrated as shown in Table \ref{tab:finetune_llff}. We compare our method EVE-NeRF with NeRF~\cite{mildenhall2021nerf}, NeX~\cite{wizadwongsa2021nex}, and NLF~\cite{suhail2022light}. Our method outperforms baselines on the average metrics. The LPIPS of our method is lower than NLF by $13.4\%$, although NLF requires larger batchsize and longer iterations of training.



\subsection{Qualitative Comparison With Na\"ive Dual Network Methods}
\label{Qualitative Comparison With naive}
As depicted in Figure \ref{naive_compare_llff}, we showcase a qualitative comparison of our approach with two other dual-branch methods on the Room and Horns scenes from the LLFF dataset. Our approach exhibits fewer artifacts and a more accurate geometric appearance. Specifically, in the Room scene, our method avoids the black floating artifacts seen in the chair and wall in the other two methods. In the Horns scene, our approach accurately reconstructs the sharp corners without causing ghosting effects. Figure \ref{naive_compare_blender} illustrates the qualitative comparison results in the Materials scene from the Blender dataset. It is evident that our method outperforms other dual-branch methods in rendering quality.

While adding the cross-attention interaction mechanism can enhance the performance of generalizable novel view synthesis, it is apparent from Figure \ref{naive_compare_fig} that the rendered novel view images still exhibit artifacts and unnatural geometry. In some cases, the reconstruction quality of certain objects may even be inferior to the na\"ive dual transformer, as observed in the upper-left part of Figure \ref{naive_compare_blender}. This could be attributed to the limitation of the cross-attention interaction mechanism in aggregating features across both epipolar and view dimensions simultaneously.

Furthermore, we individually visualized the rendering results of each branch within the Na\"ive Dual Transformer, as depicted in Figure \ref{Branch Rendering Results}. It was observed that the second branch based on the epipolar transformer produced blurry rendering results. This is likely due to the absence of geometric priors, as interacting with epipolar information first can make it challenging for the model to acquire the geometry of objects.
Therefore, aggregating view-epipolar feature na\"ively may cause pattern conflict between view dimension and epipolar dimension. Instead of na\"ive feature aggregation, the dual network architecture of EVE-NeRF aims to compensate for the inadequacies in the first branch's interaction with information in the epipolar or view dimensions, providing the appearance continuity prior and the geometry consistency priors.


\section{Limitation}
Although our approach achieves superior performance in cross-scene novel view synthesis, it takes about 3 minutes to render a novel view image with a resolution of $1008 \times 756$, which is much longer than the vanilla scene-specific NeRF approach~\cite{mildenhall2021nerf,hedman2021baking,chen2023mobilenerf}. Nevertheless, we must admit that the simultaneous achievement of high-quality, real-time, and generalizable rendering poses a considerable challenge. In light of this, we posit that a potential avenue for further exploration is optimizing the speed of generalizable NeRF. 


\end{document}